\documentclass{article}

\usepackage{microtype}
\usepackage{graphicx}
\usepackage{subfigure}
\usepackage{booktabs} 

\usepackage{hyperref}

\usepackage[accepted]{icml2025}

\usepackage{amsmath}
\usepackage{amssymb}
\usepackage{mathtools}
\usepackage{amsthm}

\usepackage[capitalize,noabbrev]{cleveref}

\theoremstyle{plain}

\theoremstyle{definition}

\theoremstyle{remark}

\usepackage[textsize=tiny]{todonotes}

\icmltitlerunning{Beyond Sensor Data: Foundation Models of Behavioral Data from Wearables Improve Health Predictions}

\begin{document}

\twocolumn[
\icmltitle{Beyond Sensor Data: Foundation Models of Behavioral Data from Wearables Improve Health Predictions}

\icmlsetsymbol{equal}{*}
\icmlsetsymbol{senior}{†}

\begin{icmlauthorlist}
\icmlauthor{Eray Erturk}{equal,yyy}
\icmlauthor{Fahad Kamran}{equal,comp}
\icmlauthor{Salar Abbaspourazad}{comp}
\icmlauthor{Sean Jewell}{comp}
\icmlauthor{Harsh Sharma}{comp}
\icmlauthor{Yujie Li}{comp}
\icmlauthor{Sinead Williamson}{comp}
\icmlauthor{Nicholas J Foti}{senior,comp}
\icmlauthor{Joseph Futoma}{senior,comp}
\end{icmlauthorlist}

\icmlaffiliation{yyy}{USC (Work done during an internship at Apple.)}
\icmlaffiliation{comp}{Apple Inc}

\icmlcorrespondingauthor{Joseph Futoma}{jfutoma@apple.com}


\vskip 0.3in
]

\printAffiliationsAndNotice{\icmlEqualContribution †Equal senior authors}  

\begin{abstract}

Wearable devices record physiological and behavioral signals that can improve health predictions. 
While foundation models are increasingly used for such predictions, they have been primarily applied to low-level sensor data, despite behavioral data often being more informative due to their alignment with physiologically relevant timescales and quantities.
We develop foundation models of such behavioral signals using over 2.5B hours of wearable data from 162K individuals, systematically optimizing architectures and tokenization strategies for this unique dataset. 
Evaluated on 57 health-related tasks, our model shows strong performance across diverse real-world applications including individual-level classification and time-varying health state prediction.
The model excels in behavior-driven tasks like sleep prediction, and improves further when combined with representations of raw sensor data.
These results underscore the importance of tailoring foundation model design to wearables and demonstrate the potential to enable new health applications.
\end{abstract}


\section{Introduction}

\begin{figure*}
    \centering
    \includegraphics[trim={0 0.25cm 0 0},clip,width=0.9\linewidth]{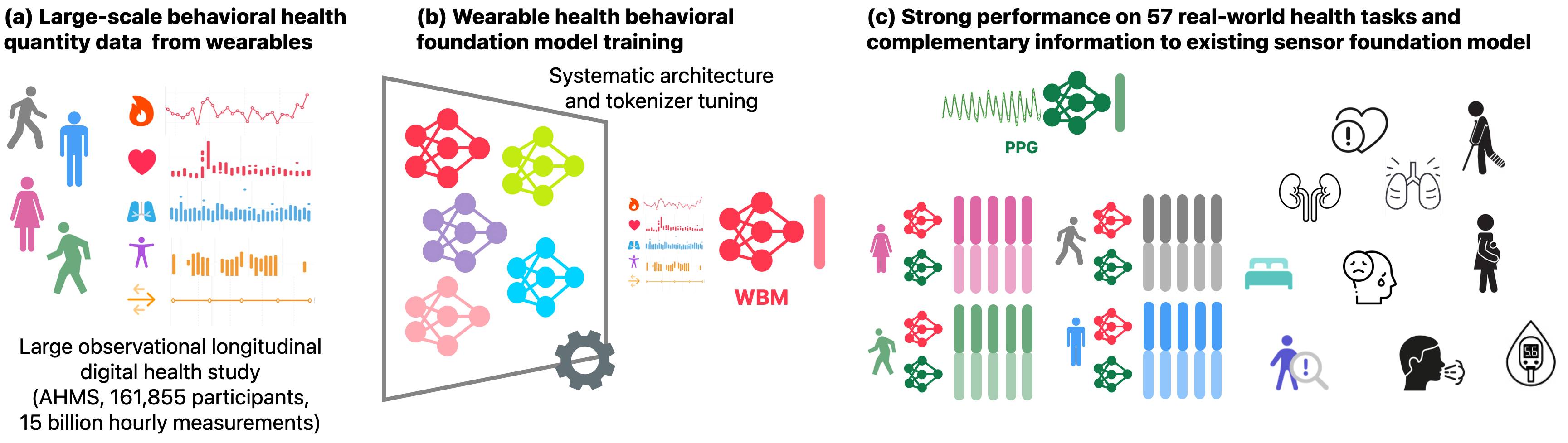}
        \vspace*{-4mm}
        \caption{An overview of our approach to solving a wide variety of health detection tasks. We use the Apple Heart and Movement Study to develop a foundation model for irregularly sampled behavioral time series data. We systematically optimize tokenizers and architectures for our dataset to build our final model, and we find that it excels in behavior-driven tasks and provides complementary information to an existing PPG foundation model.}
    \label{fig:overview}
\end{figure*}

Consumer wearables, such as smartwatches and fitness trackers, provide rich information across diverse health domains, including cardiovascular health \citep{moshawrab2023smart}, sleep \citep{bianchi2018sleep}, diabetes \citep{ahmed2022overview}, mental health \citep{gomes2023survey}, respiratory health \citep{channa2021rise}, reproductive health \citep{lyzwinski2024innovative}, and more.
A common type of prediction problem in wearables is detection, where a set of physiological, behavioral, and/or sensor data are used to predict information about a person's current health state.
An important aspect of health monitoring is detecting a static health state -- for instance, whether someone has a history of smoking, has a past diagnosis of hypertension, or is on a beta-blocker.
Another crucial problem is detecting a transient health state, such as the quality of someone's sleep or whether someone is currently pregnant.
A key property of the data required for these predictions is that they are typically at the temporal resolution of human behavior (e.g., days and weeks) rather than at the lower-level time scales (e.g., seconds) at which raw sensor data is collected from wearables.

Though a majority of past work has considered modeling low-level sensor data (or simple features thereof) \cite{abbaspourazad2024wearable,narayanswamy2024scaling,apple_ppg_FM_2024,yuan2024self}, higher-level behavioral information from wearables such as physical activity, cardiovascular fitness, and mobility metrics, are the natural data type to help solve these detection tasks. 
Unlike raw sensors, these higher-level behavioral metrics are calculated using carefully validated algorithms derived from the raw sensors. 
These metrics are intentionally chosen by experts to align with physiologically relevant quantities and health states. Importantly, these data are sensitive to an individual's behaviors, rather than being driven purely by physiology. These characteristics make behavioral data particularly promising for such health detection tasks. 
For example, mobility metrics that characterize walking gait and overall activity levels may be important behavioral factors to help detect a changing health state such as pregnancy.

However, behavioral data present unique challenges, including irregular sampling, missing data, and variable sampling rates across features, requiring specialized approaches for modeling.
Moreover, it is common to have massive quantities of unlabeled, real-world wearable data, but only have higher-quality labels on a small curated subset. This motivates the need for approaches to learn useful representations of these behavioral wearable data in the absence of labels.

As such, these data naturally align with foundation models \citep{bommasani2021opportunities}, where pre-training on large, diverse datasets enables transfer learning to new tasks with limited labels.
Foundation models can capture complex, nonlinear interactions in wearable data, reducing the need for task-specific models and addressing the challenge of small, often weakly labeled datasets, a common limitation in wearable health research.
Despite the growing interest in foundation models for wearable data, prior work has primarily focused on modeling low-level biosignals, such as photoplethysmogram (PPG), electrocardiogram (ECG), and accelerometer data \citep{apple_ppg_FM_2024, yuan2024self, abbaspourazad2024wearable}, or simple extracted features from them \citep{narayanswamy2024scaling}.
These signals, while valuable, are not consistently available throughout the day, limiting their applicability across diverse health detection problems.
In contrast, behavioral data provide complementary insights by capturing broader patterns of health and activity.
Integrating behavioral data with raw sensor signals offers a holistic view of an individual's health, enabling models to generalize across a wider range of detection problems.

In this work, to realize the goal of accurate health detection we develop a wearable health behavior foundation model (WBM).  
WBM is trained on behavioral data from wearables, using 162K participants with over 15 billion hourly measurements from the Apple Heart and Movement Study \citep{macrae_apple_2021,shapiro_pulse_2023}.
Our model development process was guided by an understanding of the unique challenges posed by behavioral data, such as irregular sampling and missingness.
We conducted a principled exploration of existing state-of-the-art approaches from past literature in wearables and time series modeling to understand and test which existing solutions would translate best to our unique data, resulting in a single performant model.   
We systematically evaluate WBM on 57 health-related detection tasks, including both existing and novel tasks that span a variety of medical domains. We demonstrate our foundation model's ability to encode important health information, outperforming a strong baseline and complementing an existing PPG foundation model.
See Figure \ref{fig:overview} for a high-level summary of our work.

Our key contributions include:
\textbf{1) Strong performance of behavioral data for health detection:} We demonstrate WBM's generalizability and real-world utility across diverse health detection tasks.
\textbf{2) Integrating behavioral and sensor data:} We show that incorporating WBM as a model of behavioral data improves upon an existing PPG foundation model across most tasks, highlighting their complementary strengths. 
\textbf{3) Developing a foundation model for wearables behavioral data:} We outline the design and development of the first large-scale foundation model for irregularly sampled behavioral time series data.

\section{Related Work}

\subsection{Foundation Models for Wearable Sensors}

Recent work has demonstrated the success of large-scale pretraining of foundation models via self-supervised learning (SSL) on low-level wearable sensor data, such as PPG, ECG, and accelerometer signals, to create general-purpose, accurate models \citep{apple_ppg_FM_2024, narayanswamy2024scaling, yuan2024self, abbaspourazad2024wearable, xu2024relcon}. 
These models are typically trained on raw sensor streams or extracted signal features from low-level sensors, leveraging SSL objectives to capture rich representations of biosignals.
Our work is complementary, focusing instead on modeling higher-level behavioral data derived from wearables, such as activity and mobility metrics, which are more computationally efficient and on more physiologically relevant timescales for many health detection tasks. 
\citet{merrill2023self} is the only prior work we are aware of that has explored SSL methods on behavioral data, but it was limited to only three variables (heart rate, step count, and a sleep/wake flag) on a small dataset of 5.2K individuals. 
In contrast, we model a more diverse set of behavioral signals derived from wearables, using a large-scale dataset with tens of millions of participant-weeks, enabling more robust and generalizable foundation models.

\subsection{Foundation Models for Time Series}
Behavioral data from wearables can be represented as a multivariate time series, but poses unique challenges compared to traditional time series, including higher rates of missing data, irregular sampling, and varying sampling frequencies across variables. 
Foundation models for time series are an active area of research, particularly for forecasting and anomaly detection tasks in domains such as finance, energy, and climate \citep{lotsa2024, timegpt2023, chronos2024, lagllaam2023}. 
These approaches generally use a decoder-only approach. However, as we are not interested in forecasting or predicting the wearable data itself, we focus on an encoder-only framework for learning meaningful representations. 

Despite the progress in time series foundation models, few methods directly address the challenges posed by wearable behavioral data.
Irregular time series modeling approaches have often been explored in the context of electronic health records and clinical time series \citep{primenet2023, mTAN2021, tuple2022,zhang2023warpformer,labach2023duett}, which do not generally capture behavioral information despite having other commonalities in terms of missingness and irregularity.
We consider approaches from this past work, conducting principled ablations to find the best model for our data by combining these ideas with other state-of-the-art sequence architectures, such as the Mamba-2 architecture \citep{gu2023mamba, dao2024transformers} which have shown promise for healthcare applications \citep{fallahpour2024ehrmamba, wornow2024context}.

\section{Data}

We use a variety of wearable data collected under informed consent from participants in the Apple Heart and Movement Study (AHMS) \citep{macrae_apple_2021,
shapiro_pulse_2023,truslow_understanding_2024}, an ongoing real-world observational cohort study exploring the relationship between cardiovascular measures, activity, and outcomes.
Sponsored by Apple Inc in partnership with the American Heart Association and Brigham and Women's Hospital, AHMS includes over 278,000 participants with up to 5 years of longitudinal data.
Participants, who are U.S. residents aged 18+, provide informed consent using the Apple Research app \citep{truslow_understanding_2024} and control the types of data they share.
The dataset comprises wearable-derived metrics from the Apple Watch and iPhone, and health outcomes from surveys and medical records. To our knowledge, this is the largest and most diverse wearable data utilized to date \citep{bycroft2018uk,denny_all_2019,merrill2023homekit2020}. 

Our work focuses on 27 interpretable HealthKit quantities\footnote{\scriptsize https://developer.apple.com/documentation/healthkit/hkquantitytype} that are calculated from lower-level sensors using validated methods. These metrics encode both physiological and behavioral information. 
Compared to modeling raw sensor data, these derived metrics are chosen by experts due to their alignment with meaningful physiological health states, making these metrics particularly suited for downstream modeling tasks.
For example, VO2Max, a measure of cardiovascular fitness estimated from outdoor workouts, reflects both behavioral patterns (e.g. exercise intensity) and cardiovascular health.
Other metrics, such as step count, primarily encode behavioral activity.


Below, we group the metrics into categories based on the types of information they capture.
Throughout this paper, we refer to these as \textit{behavioral data} or \textit{behavioral health quantities}. However, for some variables, such as cardiovascular and vital signs, the measurements reflect a combination of underlying physiology and behavior.
For example, exertion (a behavior) increases heart rate, but heart rate itself is also a direct physiological signal.
A complete description of each variable and its availability is provided in Appendix {Table \ref{appendix:variable_list}.


\textbf{Activity (8 variables):} active energy (estimated calories burned), basal energy, step count (phone and watch), exercise time, standing time, and flights climbed (phone and watch).

\textbf{Cardiovascular (4 variables):} resting heart rate, walking heart rate average, heart rate, and heart rate variability.

\textbf{Vitals (3 variables):} respiratory rate (overnight only), blood oxygen, and wrist temperature (overnight).

\textbf{Gait / mobility (8 variables):} walking metrics (speed, step length, double support percentage, asymmetry percentage, and steadiness score), stair ascent/descent speed, and fall count.

\textbf{Body Measurements (2 variables):} body mass and BMI.

\textbf{Cardiovascular Fitness / Functional Capacity (2 variables):} VO2max and six minute walk distance, both clinically validated measures of fitness and capacity.


While AHMS is one of the largest wearable research datasets collected to date, the study cohort is not fully representative of the broader U.S. population.
Participants are iPhone and Apple Watch users who voluntarily opted into a digital research study, which may introduce selection bias related to socioeconomic status, technology access, and health engagement. 
Additionally, despite the large cohort size, certain demographic groups, such as women, older adults, and racial and ethnic minorities remain underrepresented.
The demographics of the cohort used to train our foundation model can be found in Appendix Table ~\ref{table:appendix_demog_details}.
These considerations are important when interpreting results and assessing generalizability. 
Nonetheless, the scale and richness of the data make it uniquely well-suited for developing foundation models of human physiology and behavior from wearable sensors, as we describe in the next section.




\section{Developing a Foundation Model}


To improve predictions on a wide variety of health detection tasks, we developed a foundation model capable of encoding an individual's health and behavioral data over a time window into a single embedding. In Figure \ref{fig:arch}, we showcase the model architecture and tokenizer that we found performed best. In the remainder of this section, we discuss the motivation and empirical analysis that led us to particular dataset and modeling choices. 

\begin{figure}
    \centering
    \includegraphics[trim={0 0.25cm 0 0},clip,width=1\linewidth]{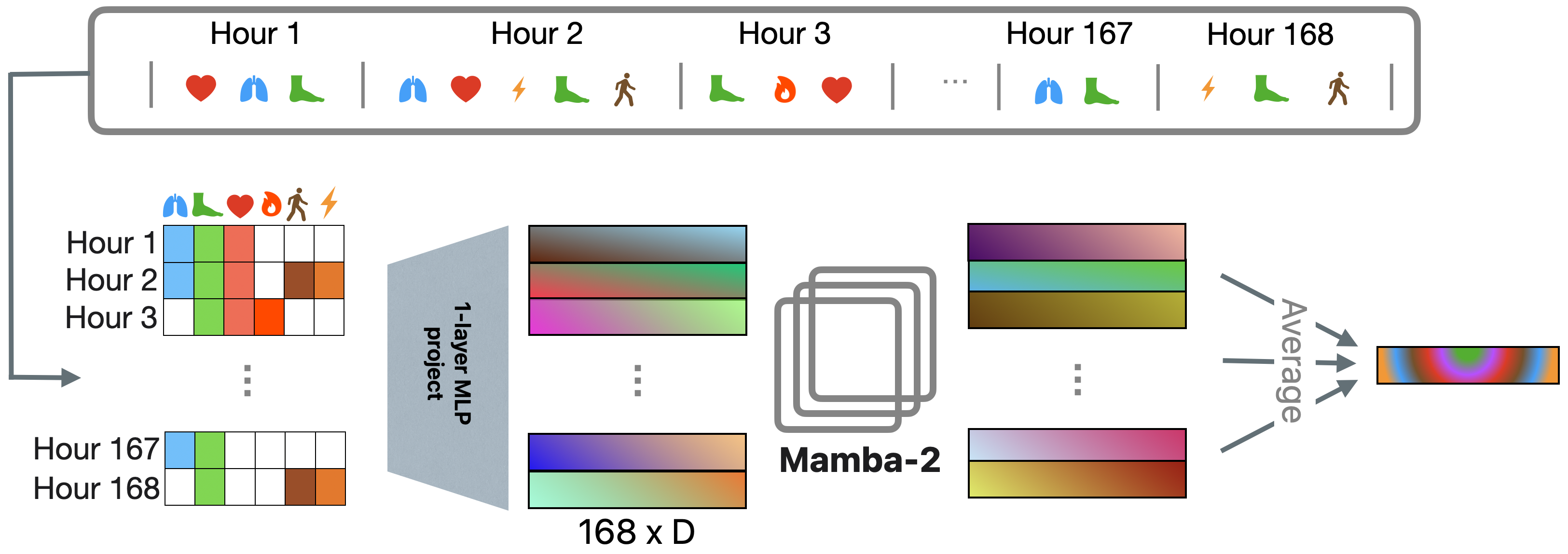}
        \vspace*{-7mm}
        \caption{The final WBM embedding pipeline for an individual's  behavioral data. The input data is irregularly sampled, both within a specific variable and across all 27 variables. We transform this into a dense matrix of weekly data, where each row is an hour of data. We also include a missingness mask (not shown) for each variable across the week, resulting in a 168 $\times$ 54 matrix. We project each hour as a patch using a single linear layer, and use the resulting patches as input to Mamba-2. Finally, we average the Mamba-2 outputs across time to create a single embedding for an individual's week of data.}
    \label{fig:arch}
\end{figure}

\subsection{Pre-training Dataset}

We created the pre-training dataset by aggregating behavioral health data at hourly intervals, aligning with our goal of detecting health states over longer time scales.
Hourly aggregation promotes consistency across variables with uneven sampling rates, and is also a common design decision in prior work \citep{lipton2016directly,xu2018raim,wang2020mimic}. 
Driven by our goal of detecting health states at a temporal resolution of human behavior, weekly windows of aggregated hourly data were chosen as model inputs, balancing prediction granularity, actionability, and scalability.

The final dataset consists of 15.14 million weeks of data ($\sim$2.5B hours) from 161,855 unique participants, filtered for adequate watch wear (see Appendix  \ref{appendix:variable_info} for details).
Data irregularity stems from variable sampling rates, missing data, and participant-specific inconsistencies.
We report detailed statistics on variable availability in Appendix \ref{appendix:variable_info} and Table \ref{appendix:variable_list}.
We use the same 80\% / 10\% / 10\% participant-level splits for training, validation, and testing for both pre-training and downstream tasks.

\subsection{Model Training}

The irregularity of our data significantly differs from typical biosignals, necessitating a comprehensive investigation into what techniques work best. Moreover, as this type of behavioral data has not been previously used to build foundation models, there is no direct prior knowledge to inform either the best methods for tokenizing variables as inputs nor the best backbone architectures. As such, we perform a principled empirical exploration of different input tokenizers and model backbones most appropriate for our data. 

\textbf{Input Tokenizers}
Understanding how to correctly represent the data is critical for handling irregularly sampled data. A tokenizer maps an individual's week of data to a sequence of vectors that can be used as input to a deep learning model. We consider three tokenizers that can be applied to our input data. We provide a brief overview in this section, and leave specific implementation details to Appendix \ref{appendix:tokenization}. 

\textit{TST.} First, we consider a dense approach for modeling a week of an individual's data inspired by the Time Series Transformer (TST) approach \citep{tsts2021}. We create a $168 \times 27$ matrix for each hour in the week and all 27 variables. For each of the 27 variables, we crudely impute missing hours with a global average value (i.e., zero as we use z-scored inputs). Though we tried more sophisticated versions of imputation (e.g., subject-level averages or weekly averages), the simpler global average resulted in the best performance in initial experiments. We also concatenate missingness indicators for each variable, resulting in a 168 $\times$ 54 feature matrix. We treat each hour of data as a patch, and use a multi-layer perceptron with one hidden layer to map each patch to an embedding vector \citep{tsts2021}. This sequence of vectors can then be used as input to a backbone deep learning model.  

\textit{mTAN.} Next, we consider using multi-time attention (mTAN) with masking to appropriately handle missing data. We use the same 168 $\times$ 54 feature matrix from TST as input to a mTAN network, akin to past work \citep{primenet2023, mTAN2021}. mTAN returns a sequence of embedding vectors that can then be used as input to a backbone deep learning model. 

\textit{Tuple.} Finally, we consider a non-dense approach that treats each observation of data as a tuple containing time (i.e., hour in week measurement was taken), variable type, and numerical value. We adapt techniques from past work which learn one-to-many mappings from scalar-valued time- and variable-values to a higher dimensional embedding, and use an embedding lookup table to map each categorical variable type to higher dimensions \citep{tuple2022,zhang2023warpformer,labach2023duett}. Adding these three embeddings results in a single input token for each hourly variable measurement. Missing values are naturally handled, as variables that were not measured at a given hour are simply omitted. The sequence of all tokens can then be used as input to a backbone deep learning model.

\textbf{Backbone Model Architectures.}
Given input tokens of the irregularly sampled data, we describe a few popular sequence-to-sequence architectures to learn meaningful representations. All models take in a sequence of embedding tokens, and return an output sequence of the same length. To obtain a single output embedding, we average across time over all output embeddings of the model. We provide a brief overview, but leave details to  Appendix \ref{appendix:architectures}. 

\textit{Self-Attention Transformer.} Given the irregularity of our input data and the strong results of Transformers across different modalities, including wearable data, we first consider using a Transformer to learn representations of the data \citep{attention2017}. Given the importance of representing temporal aspects of the signals, we pay special attention to positional encodings. First, we consider standard self-attention techniques with learnable position encodings. 

\textit{Rotary Transformer.} Absolute positions may matter less than relative positions when working with temporal data such as wearable signals. Hence, we also explore the utility of using relative position encodings. In particular, we consider Rotary Position Embeddings (RoPE), due to their strong performance compared to other variants, and strong theoretical properties that show their ability to flexibly encode both absolute and relative encodings \citep{shaw2018self,ren2021rapt,dufter2022position,su2024roformer}.

\textit{Mamba-2.} Continuous time state-space models are natural choices for handling irregularly sampled time series data due to their learned discretization step sizes.
Specifically, we consider the recent strand of selective state-space models known as Mamba, as they are efficient, more flexible than classical state-space models, and have been shown to be competitive with Transformers for language modeling \citep{gu2023mamba,dao2024transformers}.
We consider using Mamba-2 as a backbone architecture to learn downstream representations. We apply a bi-directional Mamba-2 architecture as in past work using state-space models for time series, allowing the representations to summarize information in both the forward and backward directions \citep{wang2024mamba,liang2024bi}. 

\textbf{Pre-Training Loss.} We use a regularized contrastive objective as our SSL pre-training loss. Similar contrastive objectives have achieved strong performance in health-related time-series signals \citep{jeong2023EBCL}, and in learning performative foundation models from wearable signals \citep{apple_ppg_FM_2024}. Due to the importance of capturing sparse but informative variables (e.g., VO2max), we do not consider a masked autoencoder pre-training framework. These techniques require the model to be able to recreate all portions of the input signal, which may both overemphasize observed variables and be unnecessarily stringent. Though such a task is useful for imputation and interpolation, these tasks are not a focus of this work. In Appendix \ref{appendix:MAE}, we show the poor performance of a masked autoencoder in initial experiments, further motivating the use of contrastive approaches in downstream health detection tasks. Though other SSL objectives exist and may be suitable, we leave an exploration of these non-contrastive approaches to future work.

We use a subject-level positive sampling contrastive objective (see Appendix \ref{appendix:loss} for further details), which has been shown to learn informative representations for many of the same health detection tasks we are interested in \citep{apple_ppg_FM_2024}. To encourage learning meaningful representations, we sample pairs of augmented segments (i.e., weeks of data) from the same individual. To create the augmented segments, we use input token dropping. Specifically, we consider dropping $p\%$ of tokens across time prior to learning representations during training. 


\vspace{-0.1in}
\subsection{Tuning and Best Model Creation}

To create a final performant foundation model, we train every combination of tokenization (Tuple, mTAN, and TST) and architecture (Self-Attention Transformer, Rotary Transformer, Mamba-2) using the regularized contrastive loss. We perform a grid search over all 9 combinations, tuning a subset of parameters for each model, including batch size, KoLeo regularization amount \citep{sablayrolles2018spreading,chen2020simple}, weight decay, and number of layers (see Appendix \ref{appendix:hyps} for details). To choose between these different models, we train and evaluate the learned embeddings on the representative downstream task of age prediction. As age is predictive of many health conditions, the task of age prediction is often used to measure the quality of learned embeddings \citep{apple_ppg_FM_2024,narayanswamy2024scaling}. 

Overall, we find that most models perform well on the downstream age task on the held-out validation set (Appendix Table \ref{table:hk_ablation}), with the combination of the simple TST tokenization and Mamba-2 backbone achieving the lowest error. We hypothesize that  TST tokenization outperformed more sophisticated tokenization methods due to the high level of noise in the data and significant aggregation done in preprocessing, such that the naive imputation of missing variables using a population global mean was sufficient. Despite the prevalence of Transformers for building foundation models for wearable data, we found that a Mamba-2 backbone resulted in the best model. Importantly, all models were given the same hyperparameter budget, and the best model was not always the largest model. Moreover, we capped the number of layers for Mamba-2 in this initial exploration to the largest Transformer that could fit in memory due to the attention calculation. Despite this, we found the TST and Mamba-2 combination consistently resulted in the best model across other downstream tasks as well (see Appendix Table \ref{table:hk_ablation_survey}). 

Given this initial finding, we perform a larger hyperparameter search using a Mamba-2 backbone with TST tokenization to create the final foundation model, which we denote WBM. Figure \ref{fig:arch} shows the full process of going from an individual's input data to an output embedding that can be used for downstream tasks. Though we are unable to release model weights and code due to the specifics of the informed consent for participants in the study, Section \ref{appendix:hyps} contains all training details and final model configuration. For more details, interested parties can reach out to the authors.


\vspace{-0.1in}
\section{Downstream Evaluation of Learned Foundation Model}

\begin{table}[!htb]
\footnotesize
\centering
\caption{Performance of embeddings on age and sex prediction with 95\% bootstrap confidence intervals in parenthesis. The combination of WBM and PPG consistently performs best.}
\vskip 0.1in
\begin{tabular}{ccc}
\hline
\textbf{Embedding}       & \textbf{Age (MAE)} & \textbf{Biological Sex (AUROC)} \\ \hline
Baseline   & 7.89 (6.66, 10.31) & 0.931 (0.928, 0.934) \\ \hline 
WBM    & 3.67 (3.63, 3.71) & \textbf{0.999 (0.999, 0.999)} \\ \hline 
PPG & 2.89 (2.86, 2.93)  & 0.997 (0.996, 0.997)\\ \hline \hline
WBM $+$ PPG & \textbf{2.46 (2.43, 2.50)} & \textbf{0.999 (0.999, 1.000)} \\ \hline
\end{tabular}
\vskip -0.1in
\label{table: demog_ppg_hk}
\end{table}

We now aim to understand whether the learned WBM embeddings encode meaningful information for a wide variety of health detection tasks. 

\vspace{-.08in}
\subsection{Downstream Tasks}

\textbf{Motivation.} The central research question driving our work is whether WBM encodes enough information to enable strong predictive performance across a wide range of health detection tasks. Prior work in the wearables space has largely focused on activity recognition and exercise detection tasks from sensor data \citep{narayanswamy2024scaling, yuan2024self} or on single disease states or conditions \citep{merrill2023self}, and do not comprehensively evaluate on a wide variety of different types of health conditions.
Given the unique nature of our large-scale observational study, we have access to a much wider variety of potential health, wellness, and medication outcomes via comprehensive longitudinal health surveys and medical records. 
This allows us to better assess the generalizability of our models to detect a larger variety of meaningful health states that might provide individuals actionable insights. 
We build on prior work that demonstrated promising performance of foundation models trained on low-level physiological signals like PPG and ECG for static health classification tasks \citep{apple_ppg_FM_2024}.
In this study, we not only adopt those same benchmarks but also introduce novel tasks involving time-varying health states. 
Our hypothesis is that WBM and PPG capture complementary aspects of health, and that combining them will yield the best predictive performance across most tasks.


    

\textbf{Experimental Set-Up.} We use linear probing (with a ridge penalty) to fit downstream models using WBM's embeddings as predictors for each downstream task. Models are fit either on a per-week or per-participant basis, as appropriate.
We use the same participant-level splits as in pre-training (80\%/10\%/10\%), combining validation and test sets to yield 80/20\% train/test splits, and apply internal cross validation to select a good penalty parameter for each linear model. We also compute 95\% bootstrap confidence intervals, and use these to compare different models.

\textbf{Baseline Models.} As a simple baseline, we use the mean and standard deviation of each of the 27 health behavior variables as a way to summarize each week of data, rather than using a learned model embedding. 
To fit participant-level models, similar to averaging week-level embeddings, we average these week-level vectors of statistics across all weeks of available data for a participant. 
For all downstream tasks except age and sex prediction, we also include relevant demographic information (i.e., age, sex, and BMI) as inputs. We denote this combination of statistics and demographics as ``Baseline''. Comparing against such simple baseline features shows the utility of foundation models over a typical simple supervised approach for our data. We also compare against a PPG foundation model (which we denote as ``PPG'') from prior work \citep{apple_ppg_FM_2024}, as they showed extremely strong performance on many downstream health detection tasks. Comparing against a PPG model allows us to examine situations where modeling behavioral data may be of more benefit than modeling lower-level sensor data. Finally, we create what we expected to be the strongest model by combining embeddings from the PPG model with WBM, to show the complementary nature of modeling behavioral data and lower-level sensor data. Specifically, we combine WBM and PPG embeddings by concatenating the two resulting embedding vectors into one single embedding vector that can be used as input for downstream models.

\begin{table*}[!htb]
\centering
\scriptsize
\caption{Performance of embeddings on time-varying health detection (AUROC) and sleep regression ($R^2$) tasks with 95\% bootstrap confidence intervals below in parenthesis. The combination of WBM and PPG consistently yields the best performance.}
\vskip 0.05in
\begin{tabular}{lcccccccc}
\hline
\multicolumn{1}{c}{\textbf{Embeddings}} & \textbf{\begin{tabular}[c]{@{}c@{}}Diabetes\\ (AUROC)\end{tabular}} & \textbf{\begin{tabular}[c]{@{}c@{}}Pregnancy\\ (AUROC)\end{tabular}} & \textbf{\begin{tabular}[c]{@{}c@{}}Infection\\ (AUROC)\end{tabular}} & \textbf{\begin{tabular}[c]{@{}c@{}}Injury\\ (AUROC)\end{tabular}} & \textbf{\begin{tabular}[c]{@{}c@{}}Sleep\\ Duration ($R^2$)\end{tabular}} & \textbf{\begin{tabular}[c]{@{}c@{}}Sleep\\ Efficiency ($R^2$)\end{tabular}} & \textbf{\begin{tabular}[c]{@{}c@{}}Deep Sleep\\ Duration ($R^2$)\end{tabular}} & \textbf{\begin{tabular}[c]{@{}c@{}}REM Sleep\\ Duration ($R^2$)\end{tabular}} \\ \hline
Baseline & \begin{tabular}[c]{@{}c@{}}0.737\\ (0.729, 0.744)\end{tabular} & \begin{tabular}[c]{@{}c@{}}0.804\\ (0.795, 0.813)\end{tabular} & \begin{tabular}[c]{@{}c@{}}0.632\\ (0.626, 0.638)\end{tabular} & \begin{tabular}[c]{@{}c@{}}0.608\\ (0.605, 0.611)\end{tabular} & \begin{tabular}[c]{@{}c@{}}0.104\\ (0.093, 0.112)\end{tabular} & \begin{tabular}[c]{@{}c@{}}0.131\\ (0.127, 0.135)\end{tabular} & \begin{tabular}[c]{@{}c@{}}0.172\\ (0.168, 0.176)\end{tabular} & \begin{tabular}[c]{@{}c@{}}0.128\\ (0.124, 0.131)\end{tabular} \\ \hline
WBM & \begin{tabular}[c]{@{}c@{}}0.765\\ (0.758, 0.772)\end{tabular} & \begin{tabular}[c]{@{}c@{}}0.864\\ (0.855, 0.873)\end{tabular} & \begin{tabular}[c]{@{}c@{}}0.749\\ (0.744, 0.755)\end{tabular} & \begin{tabular}[c]{@{}c@{}}0.680\\ (0.677, 0.683)\end{tabular} & \begin{tabular}[c]{@{}c@{}}0.590\\ (0.587, 0.594)\end{tabular} & \begin{tabular}[c]{@{}c@{}}0.424\\ (0.419, 0.429)\end{tabular} & \begin{tabular}[c]{@{}c@{}}0.266\\ (0.261, 0.270)\end{tabular} & \begin{tabular}[c]{@{}c@{}}0.326\\ (0.322, 0.331)\end{tabular} \\ \hline
PPG & \textbf{\begin{tabular}[c]{@{}c@{}}0.829\\ (0.823, 0.836)\end{tabular}} & \begin{tabular}[c]{@{}c@{}}0.873\\ (0.865, 0.882)\end{tabular} & \begin{tabular}[c]{@{}c@{}}0.730\\ (0.725, 0.735)\end{tabular} & \begin{tabular}[c]{@{}c@{}}0.673\\ (0.671, 0.676)\end{tabular} & \begin{tabular}[c]{@{}c@{}}0.110\\ (0.106, 0.114)\end{tabular} & \begin{tabular}[c]{@{}c@{}}0.182\\ (0.178, 0.186)\end{tabular} & \begin{tabular}[c]{@{}c@{}}0.327\\ (0.323, 0.331)\end{tabular} & \begin{tabular}[c]{@{}c@{}}0.230\\ (0.226, 0.235)\end{tabular} \\ \hline \hline
WBM + PPG & \begin{tabular}[c]{@{}c@{}}0.828\\ (0.822, 0.834)\end{tabular} & \textbf{\begin{tabular}[c]{@{}c@{}}0.921\\ (0.914, 0.928)\end{tabular}} & \textbf{\begin{tabular}[c]{@{}c@{}}0.763\\ (0.757, 0.768)\end{tabular}} & \textbf{\begin{tabular}[c]{@{}c@{}}0.688\\ (0.685, 0.691)\end{tabular}} & \textbf{\begin{tabular}[c]{@{}c@{}}0.601\\ (0.598, 0.605)\end{tabular}} & \textbf{\begin{tabular}[c]{@{}c@{}}0.438\\ (0.433, 0.443)\end{tabular}} & \textbf{\begin{tabular}[c]{@{}c@{}}0.383\\ (0.379, 0.387)\end{tabular}} & \textbf{\begin{tabular}[c]{@{}c@{}}0.393\\ (0.388, 0.397)\end{tabular}}
\end{tabular}
\label{table:combined_results}
\end{table*}

\begin{figure*}[!htb]
    \centering
    \includegraphics[width=.9\linewidth]{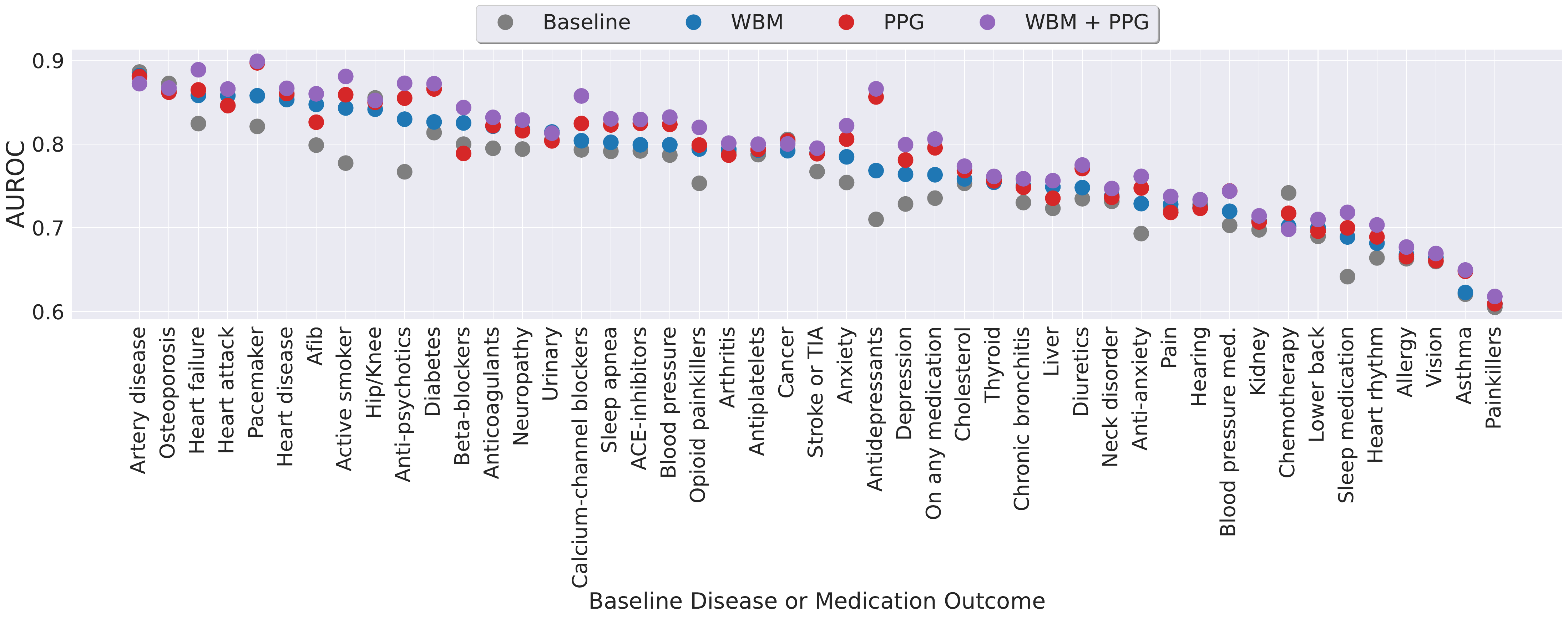}
    \vspace*{-4mm}
    \caption{Results for predicting baseline medical history and medications (AUROC). The learned representations from WBM achieve strong performance and consistently outperform the baseline. Furthermore, the combination of WBM and PPG performs best in almost every task, demonstrating the complementarity of behavioral data and sensor data for health detection tasks. See Appendix \textbf{Table \ref{table:hk_survey_full_results}} for the exact numbers for each model and task.}
    \label{fig:survey_baseline_results}
\end{figure*}

\textbf{Task Details.} To comprehensively evaluate our foundation model's ability to enable diverse health detection applications, we selected downstream tasks that fall into two broad categories: inter-subject (i.e., static) and intra-subject (i.e., dynamic or time-varying) detection tasks.

Inter-subject tasks involve predicting stable attributes or baseline conditions across individuals, such as demographic variables or medical history at the time of enrollment.
These tasks allow us to benchmark performance against prior foundation modeling work in wearables \citep{apple_ppg_FM_2024}, ensuring comparability and reproducibility. 
In particular, we include age and biological sex prediction, as they serve both as widely-used proxies for general health status and as sanity checks for representation quality, since well-trained embeddings should reliably encode this information.
While there is some potential for label leakage, given that certain WBM input variables are derived from algorithms that explicitly incorporate age and sex, we consider this acceptable in context, as these tasks are primarily used for validation rather than as potential deployment targets.
Furthermore, inter-subject predictions may be useful in real-world settings for digital health personalization or screening applications, where quickly inferring baseline characteristics without requiring detailed input from users can improve usability and accessibility.

Intra-subject tasks, by contrast, are designed to test whether our model can track meaningful within-person variation over time, capturing transient or emerging health states.
These tasks are unique to our work and highlight an important use case for wearables: continuous monitoring and detection of dynamic conditions. 
For example, we frame respiratory infection detection as a weekly classification problem, where the model must distinguish healthy vs. sick periods using historical behavior and physiology data. 
Other intra-subject tasks include detecting weekly sleep quality metrics, diabetes status via HbA1c labs, self-reported pregnancy, and injury states. 
These tasks probe whether the learned representations are sensitive enough to detect changes in health trajectories, a core capability for preventive and personalized digital health technologies.

Together, this diverse task suite enables us to evaluate both the generalizability of our model across individuals and its temporal sensitivity within individuals.
It spans multiple health domains and leverages both behavioral and physiological modalities, providing a rigorous assessment of WBM's utility as a foundation model for real-world health detection.

Full details on preprocessing and task setup can be found in Appendix \ref{appendix:task_details}.


\subsection{Results and Key Takeaways}
Table \ref{table: demog_ppg_hk} contains the demographics prediction results, Figure \ref{fig:survey_baseline_results} contains the baseline medical history and medication results, and Table \ref{table:combined_results} contains the intra-subject results. Additional results within demographic subgroups for a select number of tasks can be found in Appendix \ref{appendix:additional_results}.

\textbf{Foundation models of behavioral data encode substantial health information.} WBM achieves strong performance across all tasks. Notably, WBM substantially outperforms the simple supervised baseline on all demographic tasks. Moreover, WBM is consistently strongly predictive of most baseline disease and medication health outcomes, and usually outperforms the simple baseline. In total, WBM outperforms the baseline in 39 out of 47 outcomes, with a median AUROC improvement of 0.017 in these outcomes and a statistically significant improvement in 30 out of 47 outcomes. In the eight outcomes in which the baseline model performs best, WBM is still competitive (median AUROC deficit of 0.008). Finally, WBM is able to achieve strong results across intra-subject tasks as well, statistically significantly outperforming the baseline model in all 8 time-varying health detection tasks. As a sanity check, we also show that WBM accurately reconstructs the weekly mean values for most of the 27 variables considered, including a majority of the sparser input variables such as mobility metrics, in the Appendix \textbf{Table \ref{table:hk_stats_embed_preds}}.

\textbf{Behavioral wearable data contains more signal than low-level sensor data for some tasks.} WBM outperforms the PPG model across several tasks, especially those where we expect behavioral information should provide extra signal. This is most apparent in the sleep tasks, where WBM results in consistent performance improvements. 
This is not surprising -- behavioral data capture information from every hour of the week, including overnight periods, where we can infer how long someone was asleep by periods of inactivity (e.g., from step count, exercise minutes, and heart rate).
However, PPG does not provide as comprehensive a view of an individual’s week, since it is only measured a few times each day.
WBM also provides gains in other tasks such as infection and injury, where behavioral data may provide additional information about an individual's activity that results in more accurate models compared to lower-level PPG sensors (e.g., gait and mobility metrics will likely change after a lower limb injury). 
WBM also outperforms PPG on 18 of the 47 baseline disease and medication outcomes, including four with statistically significant differences.
Notably, WBM achieves better performance in predicting beta blocker use, which is consistent with its ability to more reliably detect heart rate reductions throughout the day, information that may be missed by the more sporadic measurements of PPG.

\textbf{Low-level sensor data outperforms behavioral data in tasks where physiological information is sufficient.} There are many tasks in which the lower-level PPG data is sufficient, and so lower-level physiology provides enough information to achieve strong performance. This is most apparent in the time-varying diabetes task, where the PPG model outperforms all others, including when combined with the behavioral data foundation model. This also occurs consistently when performing subject-level classification of baseline disease or medication outcomes (29 out of 47 outcomes with a median performance gain of .017 and 18 significant performance improvements), as the aggregation of every PPG over the study is sufficient for detecting many baseline outcomes. This is clear when looking at the outcome of antidepressants, where PPG excels as shown in past work \citep{apple_ppg_FM_2024}, and behavioral data provides only marginal improvements.

\textbf{Models of behavioral data provide complementary information to models of low-level sensor data.} Finally, we see that across most tasks, the combination of embeddings of WBM and the PPG model result in the most accurate models. The combination achieves the best age prediction performance across all models considered, clearly outperforming either model in isolation. This trend also holds for participant-level baseline disease and medication outcome prediction, where the combination of both modalities almost always improves upon the best single modality. In total, the combination performs best in 42 out of 47 outcomes, with a median AUROC improvement of 0.009 in these outcomes and with many results being statistically significant (38 over WBM and 33 over PPG). Combining the behavioral data with PPG results in substantial gains over PPG alone in many tasks, most notably in predicting Afib (AUROC gain of 0.034),  beta-blockers (AUROC gain of 0.055), and calcium-channel blockers (AUROC gain of 0.033). Finally, the combination of modalities remains the best model in every time-varying health state prediction task except diabetes, where PPG alone was sufficient. The combination is remarkably performant for pregnancy prediction, where the combination of both data types achieves an AUROC over 0.9 and provides a relative improvement of about 0.05 over each modality alone. Pregnancy results in substantial changes in both the underlying physiology of an individual, as tracked by raw sensors, and substantial changes to an individual's behavior, as measured by derived metrics like exercise minutes, step counts, and gait. Hence, this task acts as a clear example of the complementary nature of modeling both types of data. Combining behavioral data with PPG also improves performance in all sleep-related tasks, and in detecting infection (AUROC gain of 0.033) and detecting injury (AUROC gain of 0.015).



%

\section{Discussion}

This work built performant predictive models for meaningful health detection tasks. We approached this problem by building WBM, a foundation model trained on the largest and most diverse wearables dataset to date. Specifically, we considered quantities that capture individuals' behaviors and are derived from low-level sensor data collected from wearables. These types of quantities typically are computed using carefully validated algorithms trained from limited bespoke data. The quantities are also known to relate to many health detection tasks. Focusing on these derived quantities circumvents a model needing to learn these relationships – which may not even be possible using the observational datasets used in this and other work.

The derived behavioral quantities that we model differ from the sensor data (or simple features thereof) considered in previous works due to irregularity, with different sampling frequencies and large amounts of missingness driven by behavior. 
As such, we performed a principled search over tokenization strategies and model architectures. 
The fact that the best-performing tokenizer was the TST approach that produced a dense input format is somewhat surprising, since we found that imputing missing values by the global mean across all subjects rather than using subject-specific means worked better. 
We hypothesize that this may be due to high levels of noise when estimating variable means for some subjects, and in future work it is worth exploring if more complex model-based methods for imputation might perform even better. 
Another surprising finding was that the bi-directional Mamba-2 architecture produced the best model for our tasks and data, given the ubiquity of the transformer architecture. 
This motivates further developments of state-space models for encoder-only foundation models of time series and sensor data. 
A key takeaway of our process is that when developing foundation models in the health space it is crucial to evaluate all approaches when using new types of data, and not to assume previous models will necessarily be the best for your problem and data.

We tested our model across a suite of tasks that are representative of real-world health detection problems – some from previous work and others novel. We demonstrated that models of behavioral data encode significant information about an individual's health, outperforming models derived from low-level sensors such as PPG in tasks where behavior is a meaningful predictor (e.g., sleep and injury detection), and complementing these models across a majority of other tasks by improving the predictive performance of the joint PPG + WBM model. These results expand the foundation modeling paradigm to derived behavioral quantities benefiting a breadth of real-world health detection tasks.

This work focused specifically on incorporating individuals' behaviors into predictive models for health detection tasks using a specific dataset and thus has limitations. We cannot say how well the specific model we developed will perform on other datasets, particularly when data is not collected with Apple devices (e.g., Apple Watch and iPhone). Another limitation is that the contrastive self-supervised loss requires defining positive and negative pairs which can be arduous and may affect the types of downstream tasks the learned representations can address. While our initial experiments with masked autoencoder training were not promising, we suspect this may be due to the high degree of noise present in behavior data, and hypothesize that joint embedding predictive architectures might perform better. Moreover, as many of our downstream task labels come from self-reported surveys, some labels may be inaccurate. 
Finally, WBM is not designed to forecast future health states from an individual's history.


If developed and deployed safely and responsibly, predictive models built on wearable data like WBM hold significant promise for clinical impact. 
By enabling continuous, non-invasive monitoring and early detection of meaningful health events, such models could support more proactive and personalized care—particularly for conditions where behavioral signals are strong early indicators.
In the future, these approaches could complement clinical decision-making, help triage patients for follow-up, or support just-in-time interventions, especially in populations where traditional healthcare access is limited. 
Realizing this potential will require careful attention to model fairness, calibration, interpretability, and robust validation across diverse cohorts.

\section*{Impact Statement}

Our work is focused on enhancing health detection tasks by modeling behavioral data from wearable devices. Given our focus on wearables, we acknowledge the potential societal impact of deploying advanced health prediction models given data from wearables. Improving these prediction models may exacerbate equity gaps between individuals with access to wearable devices to those without. It remains essential to consider ways to democratize the benefits and findings of our work to other forms of data and devices that ensure equity across different groups. Moreover, as our ultimate goal is to perform predictions that may inform health-related decisions, our work has potential for positive societal consequences. It is important to further evaluate our models in terms of actionability to best realize this goal.

\subsubsection*{Acknowledgments}
We would like to thank participants in the Apple Heart and Movement Study, Calum MacRae, and study staff at The Brigham and Women’s Hospital, a Harvard affiliate, without whom this work would not have been possible. 
We are grateful to Anshuman Mishra and Valerie Marchenko for important contributions to the data processing, training, and evaluation infrastructure.
We also thank Andy Miller, Guillermo Sapiro, Ian Shapiro, Jen Block, and the anonymous reviewers for providing valuable feedback, and Lindsay Hislop, Eduardo Martinez Montes and Laura Rhodes for publication coordination help.

\newpage
\bibliography{refs}

\begin{thebibliography}{54}
\providecommand{\natexlab}[1]{#1}
\providecommand{\url}[1]{\texttt{#1}}
\expandafter\ifx\csname urlstyle\endcsname\relax
  \providecommand{\doi}[1]{doi: #1}\else
  \providecommand{\doi}{doi: \begingroup \urlstyle{rm}\Url}\fi

\bibitem[Abbaspourazad et~al.(2024{\natexlab{a}})Abbaspourazad, Elachqar, Miller, Emrani, Nallasamy, and Shapiro]{apple_ppg_FM_2024}
Abbaspourazad, S., Elachqar, O., Miller, A., Emrani, S., Nallasamy, U., and Shapiro, I.
\newblock Large-scale training of foundation models for wearable biosignals.
\newblock In \emph{International Conference on Learning Representations}, 2024{\natexlab{a}}.

\bibitem[Abbaspourazad et~al.(2024{\natexlab{b}})Abbaspourazad, Mishra, Futoma, Miller, and Shapiro]{abbaspourazad2024wearable}
Abbaspourazad, S., Mishra, A., Futoma, J., Miller, A.~C., and Shapiro, I.
\newblock Wearable accelerometer foundation models for health via knowledge distillation.
\newblock \emph{arXiv preprint arXiv:2412.11276}, 2024{\natexlab{b}}.

\bibitem[Ahmed et~al.(2022)Ahmed, Aziz, Abd-Alrazaq, Farooq, and Sheikh]{ahmed2022overview}
Ahmed, A., Aziz, S., Abd-Alrazaq, A., Farooq, F., and Sheikh, J.
\newblock Overview of artificial intelligence--driven wearable devices for diabetes: scoping review.
\newblock \emph{Journal of Medical Internet Research}, 24\penalty0 (8):\penalty0 e36010, 2022.

\bibitem[Ansari et~al.(2024)Ansari, Stella, Turkmen, Zhang, Mercado, Shen, Shchur, Rangapuram, Arango, Kapoor, et~al.]{chronos2024}
Ansari, A.~F., Stella, L., Turkmen, C., Zhang, X., Mercado, P., Shen, H., Shchur, O., Rangapuram, S.~S., Arango, S.~P., Kapoor, S., et~al.
\newblock Chronos: Learning the language of time series.
\newblock \emph{arXiv preprint arXiv:2403.07815}, 2024.

\bibitem[Apple(2023)]{apple2023sleep}
Apple.
\newblock Estimating sleep stages from {A}pple watch.
\newblock \href{https://www.apple.com/healthcare/docs/site/Estimating_Sleep_Stages_from_Apple_Watch_Sept_2023.pdf}{link}, September 2023.

\bibitem[Assran et~al.(2023)Assran, Duval, Misra, Bojanowski, Vincent, Rabbat, LeCun, and Ballas]{assran2023self}
Assran, M., Duval, Q., Misra, I., Bojanowski, P., Vincent, P., Rabbat, M., LeCun, Y., and Ballas, N.
\newblock Self-supervised learning from images with a joint-embedding predictive architecture.
\newblock In \emph{Proceedings of the IEEE/CVF Conference on Computer Vision and Pattern Recognition}, pp.\  15619--15629, 2023.

\bibitem[Bianchi(2018)]{bianchi2018sleep}
Bianchi, M.~T.
\newblock Sleep devices: wearables and nearables, informational and interventional, consumer and clinical.
\newblock \emph{Metabolism}, 84:\penalty0 99--108, 2018.

\bibitem[Bommasani et~al.(2021)Bommasani, Hudson, Adeli, Altman, Arora, von Arx, Bernstein, Bohg, Bosselut, Brunskill, et~al.]{bommasani2021opportunities}
Bommasani, R., Hudson, D.~A., Adeli, E., Altman, R., Arora, S., von Arx, S., Bernstein, M.~S., Bohg, J., Bosselut, A., Brunskill, E., et~al.
\newblock On the opportunities and risks of foundation models.
\newblock \emph{arXiv preprint arXiv:2108.07258}, 2021.

\bibitem[Bycroft et~al.(2018)Bycroft, Freeman, Petkova, Band, Elliott, Sharp, Motyer, Vukcevic, Delaneau, O’Connell, et~al.]{bycroft2018uk}
Bycroft, C., Freeman, C., Petkova, D., Band, G., Elliott, L.~T., Sharp, K., Motyer, A., Vukcevic, D., Delaneau, O., O’Connell, J., et~al.
\newblock The {UK} biobank resource with deep phenotyping and genomic data.
\newblock \emph{Nature}, 562\penalty0 (7726):\penalty0 203--209, 2018.

\bibitem[Channa et~al.(2021)Channa, Popescu, Skibinska, and Burget]{channa2021rise}
Channa, A., Popescu, N., Skibinska, J., and Burget, R.
\newblock The rise of wearable devices during the {COVID}-19 pandemic: A systematic review.
\newblock \emph{Sensors}, 21\penalty0 (17):\penalty0 5787, 2021.

\bibitem[Chen et~al.(2020)Chen, Kornblith, Norouzi, and Hinton]{chen2020simple}
Chen, T., Kornblith, S., Norouzi, M., and Hinton, G.
\newblock A simple framework for contrastive learning of visual representations.
\newblock In \emph{International Conference on Machine Learning}, pp.\  1597--1607. PMLR, 2020.

\bibitem[Chowdhury et~al.(2023)Chowdhury, Li, Zhang, Hong, Gupta, and Shang]{primenet2023}
Chowdhury, R.~R., Li, J., Zhang, X., Hong, D., Gupta, R.~K., and Shang, J.
\newblock Primenet: Pre-training for irregular multivariate time series.
\newblock In \emph{Proceedings of the AAAI Conference on Artificial Intelligence}, volume~37, pp.\  7184--7192, 2023.

\bibitem[Dao \& Gu(2024)Dao and Gu]{dao2024transformers}
Dao, T. and Gu, A.
\newblock Transformers are {SSM}s: Generalized models and efficient algorithms through structured state space duality.
\newblock \emph{arXiv preprint arXiv:2405.21060}, 2024.

\bibitem[Denny et~al.(2019)Denny, Rutter, Goldstein, Philippakis, Smoller, Jenkins, Dishman, and {The All of Us Research Program Investigators}]{denny_all_2019}
Denny, J.~C., Rutter, J.~L., Goldstein, D.~B., Philippakis, A., Smoller, J.~W., Jenkins, G., Dishman, E., and {The All of Us Research Program Investigators}.
\newblock The “{All} of {Us}” {Research} {Program}.
\newblock \emph{New England Journal of Medicine}, 381\penalty0 (7):\penalty0 668--676, August 2019.
\newblock ISSN 0028-4793.
\newblock \doi{10.1056/NEJMsr1809937}.
\newblock URL \url{https://www.nejm.org/doi/10.1056/NEJMsr1809937}.
\newblock Publisher: Massachusetts Medical Society.

\bibitem[Dufter et~al.(2022)Dufter, Schmitt, and Sch{\"u}tze]{dufter2022position}
Dufter, P., Schmitt, M., and Sch{\"u}tze, H.
\newblock Position information in transformers: An overview.
\newblock \emph{Computational Linguistics}, 48\penalty0 (3):\penalty0 733--763, 2022.

\bibitem[Fallahpour et~al.(2024)Fallahpour, Alinoori, Afkanpour, and Krishnan]{fallahpour2024ehrmamba}
Fallahpour, A., Alinoori, M., Afkanpour, A., and Krishnan, A.
\newblock Ehrmamba: Towards generalizable and scalable foundation models for electronic health records.
\newblock \emph{arXiv preprint arXiv:2405.14567}, 2024.

\bibitem[Garza \& Mergenthaler-Canseco(2023)Garza and Mergenthaler-Canseco]{timegpt2023}
Garza, A. and Mergenthaler-Canseco, M.
\newblock Timegpt-1.
\newblock \emph{arXiv preprint arXiv:2310.03589}, 2023.

\bibitem[Gomes et~al.(2023)Gomes, Pato, Lourenco, and Datia]{gomes2023survey}
Gomes, N., Pato, M., Lourenco, A.~R., and Datia, N.
\newblock A survey on wearable sensors for mental health monitoring.
\newblock \emph{Sensors}, 23\penalty0 (3):\penalty0 1330, 2023.

\bibitem[Gu \& Dao(2023)Gu and Dao]{gu2023mamba}
Gu, A. and Dao, T.
\newblock Mamba: Linear-time sequence modeling with selective state spaces.
\newblock \emph{arXiv preprint arXiv:2312.00752}, 2023.

\bibitem[Jeong et~al.(2023)Jeong, Oufattole, Balagopalan, Mcdermott, Chandak, Ghassemi, and Stultz]{jeong2023EBCL}
Jeong, H., Oufattole, N., Balagopalan, A., Mcdermott, M., Chandak, P., Ghassemi, M., and Stultz, C.
\newblock Event-based contrastive learning for medical time series.
\newblock 2023.

\bibitem[Labach et~al.(2023)Labach, Pokhrel, Huang, Zuberi, Yi, Volkovs, Poutanen, and Krishnan]{labach2023duett}
Labach, A., Pokhrel, A., Huang, X.~S., Zuberi, S., Yi, S.~E., Volkovs, M., Poutanen, T., and Krishnan, R.~G.
\newblock Duett: dual event time transformer for electronic health records.
\newblock In \emph{Machine Learning for Healthcare Conference}, 2023.

\bibitem[Liang et~al.(2024)Liang, Jiang, Sun, and Lu]{liang2024bi}
Liang, A., Jiang, X., Sun, Y., and Lu, C.
\newblock Bi-mamba4ts: Bidirectional mamba for time series forecasting.
\newblock \emph{arXiv preprint arXiv:2404.15772}, 2024.

\bibitem[Lipton et~al.(2016)Lipton, Kale, and Wetzel]{lipton2016directly}
Lipton, Z.~C., Kale, D., and Wetzel, R.
\newblock Directly modeling missing data in sequences with {RNN}s: Improved classification of clinical time series.
\newblock In \emph{Machine Learning for Healthcare Conference}, pp.\  253--270. PMLR, 2016.

\bibitem[Liu et~al.(2021)Liu, Dai, So, and Le]{liu2021pay}
Liu, H., Dai, Z., So, D., and Le, Q.~V.
\newblock Pay attention to mlps.
\newblock \emph{Advances in Neural Information Processing Systems}, 34:\penalty0 9204--9215, 2021.

\bibitem[Lyzwinski et~al.(2024)Lyzwinski, Elgendi, and Menon]{lyzwinski2024innovative}
Lyzwinski, L., Elgendi, M., and Menon, C.
\newblock Innovative approaches to menstruation and fertility tracking using wearable reproductive health technology: systematic review.
\newblock \emph{Journal of Medical Internet Research}, 26:\penalty0 e45139, 2024.

\bibitem[MacRae(2021)]{macrae_apple_2021}
MacRae, C.~A.
\newblock Apple {Heart} \& {Movement} {Study}.
\newblock Clinical trial registration NCT04198194, clinicaltrials.gov, July 2021.
\newblock URL \url{https://clinicaltrials.gov/study/NCT04198194}.
\newblock submitted: November 13, 2019.

\bibitem[Merrill \& Althoff(2023)Merrill and Althoff]{merrill2023self}
Merrill, M.~A. and Althoff, T.
\newblock Self-supervised pretraining and transfer learning enable flu and {COVID}-19 predictions in small mobile sensing datasets.
\newblock In \emph{Conference on Health, Inference, and Learning}, pp.\  191--206. PMLR, 2023.

\bibitem[Merrill et~al.(2023)Merrill, Safranchik, Kolbeinsson, Gade, Ramirez, Schmidt, Foschini, and Althoff]{merrill2023homekit2020}
Merrill, M.~A., Safranchik, E., Kolbeinsson, A., Gade, P., Ramirez, E., Schmidt, L., Foschini, L., and Althoff, T.
\newblock Homekit2020: A benchmark for time series classification on a large mobile sensing dataset with laboratory tested ground truth of influenza infections.
\newblock In \emph{Conference on Health, Inference, and Learning}, 2023.

\bibitem[Mishra et~al.(2024)Mishra, Park, Shapiro, Fisher-Colbrie, Baird, Suharwardy, Zhang, Jukic, and Curry]{mishra2024trends}
Mishra, A., Park, J., Shapiro, I., Fisher-Colbrie, T., Baird, D.~D., Suharwardy, S., Zhang, S., Jukic, A. M.~Z., and Curry, C.~L.
\newblock Trends in sensor-based health metrics during and after pregnancy: descriptive data from the apple women's health study.
\newblock \emph{AJOG Global Reports}, 4\penalty0 (4):\penalty0 100388, 2024.

\bibitem[Moshawrab et~al.(2023)Moshawrab, Adda, Bouzouane, Ibrahim, and Raad]{moshawrab2023smart}
Moshawrab, M., Adda, M., Bouzouane, A., Ibrahim, H., and Raad, A.
\newblock Smart wearables for the detection of cardiovascular diseases: a systematic literature review.
\newblock \emph{Sensors}, 23\penalty0 (2):\penalty0 828, 2023.

\bibitem[Narayanswamy et~al.(2024)Narayanswamy, Liu, Ayush, Yang, Xu, Liao, Garrison, Tailor, Sunshine, Liu, et~al.]{narayanswamy2024scaling}
Narayanswamy, G., Liu, X., Ayush, K., Yang, Y., Xu, X., Liao, S., Garrison, J., Tailor, S., Sunshine, J., Liu, Y., et~al.
\newblock Scaling wearable foundation models.
\newblock \emph{arXiv preprint arXiv:2410.13638}, 2024.

\bibitem[Nestor et~al.(2023)Nestor, Hunter, Kainkaryam, Drysdale, Inglis, Shapiro, Nagaraj, Ghassemi, Foschini, and Goldenberg]{nestor2023machine}
Nestor, B., Hunter, J., Kainkaryam, R., Drysdale, E., Inglis, J.~B., Shapiro, A., Nagaraj, S., Ghassemi, M., Foschini, L., and Goldenberg, A.
\newblock Machine learning {COVID}-19 detection from wearables.
\newblock \emph{The Lancet Digital Health}, 5\penalty0 (4):\penalty0 e182--e184, 2023.

\bibitem[Ramachandran et~al.(2017)Ramachandran, Zoph, and Le]{ramachandran2017swish}
Ramachandran, P., Zoph, B., and Le, Q.~V.
\newblock Swish: a self-gated activation function.
\newblock \emph{arXiv preprint arXiv:1710.05941}, 7\penalty0 (1):\penalty0 5, 2017.

\bibitem[Rasul et~al.(2023)Rasul, Ashok, Williams, Khorasani, Adamopoulos, Bhagwatkar, Bilo{\v{s}}, Ghonia, Hassen, Schneider, et~al.]{lagllaam2023}
Rasul, K., Ashok, A., Williams, A.~R., Khorasani, A., Adamopoulos, G., Bhagwatkar, R., Bilo{\v{s}}, M., Ghonia, H., Hassen, N.~V., Schneider, A., et~al.
\newblock Lag-llama: Towards foundation models for time series forecasting.
\newblock \emph{arXiv preprint arXiv:2310.08278}, 2023.

\bibitem[Ren et~al.(2021)Ren, Wang, Zhao, and Wu]{ren2021rapt}
Ren, H., Wang, J., Zhao, W.~X., and Wu, N.
\newblock Rapt: Pre-training of time-aware transformer for learning robust healthcare representation.
\newblock In \emph{Proceedings of the 27th ACM SIGKDD Conference on Knowledge Discovery \& Data Mining}, 2021.

\bibitem[Sablayrolles et~al.(2018)Sablayrolles, Douze, Schmid, and J{\'e}gou]{sablayrolles2018spreading}
Sablayrolles, A., Douze, M., Schmid, C., and J{\'e}gou, H.
\newblock Spreading vectors for similarity search.
\newblock \emph{arXiv preprint arXiv:1806.03198}, 2018.

\bibitem[Shapiro et~al.(2023)Shapiro, Stein, MacRae, and O’Reilly]{shapiro_pulse_2023}
Shapiro, I., Stein, J., MacRae, C., and O’Reilly, M.
\newblock Pulse oximetry values from 33,080 participants in the {Apple} {Heart} \& {Movement} {Study}.
\newblock \emph{npj Digital Medicine}, 6\penalty0 (1):\penalty0 1--13, July 2023.
\newblock ISSN 2398-6352.
\newblock \doi{10.1038/s41746-023-00851-6}.
\newblock URL \url{https://www.nature.com/articles/s41746-023-00851-6}.
\newblock Number: 1 Publisher: Nature Publishing Group.

\bibitem[Shaw et~al.(2018)Shaw, Uszkoreit, and Vaswani]{shaw2018self}
Shaw, P., Uszkoreit, J., and Vaswani, A.
\newblock Self-attention with relative position representations.
\newblock \emph{arXiv preprint arXiv:1803.02155}, 2018.

\bibitem[Shukla \& Marlin(2021)Shukla and Marlin]{mTAN2021}
Shukla, S.~N. and Marlin, B.
\newblock Multi-time attention networks for irregularly sampled time series.
\newblock In \emph{International Conference on Learning Representations}, 2021.

\bibitem[Su et~al.(2024)Su, Ahmed, Lu, Pan, Bo, and Liu]{su2024roformer}
Su, J., Ahmed, M., Lu, Y., Pan, S., Bo, W., and Liu, Y.
\newblock Roformer: Enhanced transformer with rotary position embedding.
\newblock \emph{Neurocomputing}, 568:\penalty0 127063, 2024.

\bibitem[Tipirneni \& Reddy(2022)Tipirneni and Reddy]{tuple2022}
Tipirneni, S. and Reddy, C.~K.
\newblock Self-supervised transformer for sparse and irregularly sampled multivariate clinical time-series.
\newblock \emph{ACM Transactions on Knowledge Discovery from Data (TKDD)}, 16\penalty0 (6):\penalty0 1--17, 2022.

\bibitem[Truslow et~al.(2024)Truslow, Spillane, Lin, Cyr, Ullal, Arnold, Huang, Rhodes, Block, Stark, Kretlow, Beatty, Werdich, Bankar, Bianchi, Shapiro, Villalpando, Ravindran, Mance, Phillips, Earl, Deo, Desai, and MacRae]{truslow_understanding_2024}
Truslow, J., Spillane, A., Lin, H., Cyr, K., Ullal, A., Arnold, E., Huang, R., Rhodes, L., Block, J., Stark, J., Kretlow, J., Beatty, A.~L., Werdich, A., Bankar, D., Bianchi, M., Shapiro, I., Villalpando, J., Ravindran, S., Mance, I., Phillips, A., Earl, J., Deo, R.~C., Desai, S.~A., and MacRae, C.~A.
\newblock Understanding activity and physiology at scale: {The} {Apple} {Heart} \& {Movement} {Study}.
\newblock \emph{npj Digital Medicine}, 7\penalty0 (1):\penalty0 1--11, September 2024.
\newblock ISSN 2398-6352.
\newblock \doi{10.1038/s41746-024-01187-5}.
\newblock URL \url{https://www.nature.com/articles/s41746-024-01187-5}.
\newblock Publisher: Nature Publishing Group.

\bibitem[Vaswani et~al.(2017)Vaswani, Shazeer, Parmar, Uszkoreit, Jones, Gomez, Kaiser, and Polosukhin]{attention2017}
Vaswani, A., Shazeer, N., Parmar, N., Uszkoreit, J., Jones, L., Gomez, A.~N., Kaiser, {\L}., and Polosukhin, I.
\newblock Attention is all you need.
\newblock In \emph{Advances in Neural Information Processing Systems}, 2017.

\bibitem[Wang et~al.(2020)Wang, McDermott, Chauhan, Ghassemi, Hughes, and Naumann]{wang2020mimic}
Wang, S., McDermott, M.~B., Chauhan, G., Ghassemi, M., Hughes, M.~C., and Naumann, T.
\newblock Mimic-extract: A data extraction, preprocessing, and representation pipeline for mimic-iii.
\newblock In \emph{Proceedings of the ACM conference on health, inference, and learning}, pp.\  222--235, 2020.

\bibitem[Wang et~al.(2024)Wang, Kong, Feng, Wang, Zhao, Wang, and Zhang]{wang2024mamba}
Wang, Z., Kong, F., Feng, S., Wang, M., Zhao, H., Wang, D., and Zhang, Y.
\newblock Is mamba effective for time series forecasting?
\newblock \emph{arXiv preprint arXiv:2403.11144}, 2024.

\bibitem[Woo et~al.(2024)Woo, Liu, Kumar, Xiong, Savarese, and Sahoo]{lotsa2024}
Woo, G., Liu, C., Kumar, A., Xiong, C., Savarese, S., and Sahoo, D.
\newblock Unified training of universal time series forecasting transformers.
\newblock \emph{arXiv preprint arXiv:2402.02592}, 2024.

\bibitem[Wornow et~al.(2024)Wornow, Bedi, Hernandez, Steinberg, Fries, R{\'e}, Koyejo, and Shah]{wornow2024context}
Wornow, M., Bedi, S., Hernandez, M. A.~F., Steinberg, E., Fries, J.~A., R{\'e}, C., Koyejo, S., and Shah, N.~H.
\newblock Context clues: Evaluating long context models for clinical prediction tasks on {EHR}s.
\newblock \emph{arXiv preprint arXiv:2412.16178}, 2024.

\bibitem[Xu et~al.(2024)Xu, Narain, Darnell, Hallgrimsson, Jeong, Forde, Fineman, Raghuram, Rehg, and Ren]{xu2024relcon}
Xu, M.~A., Narain, J., Darnell, G., Hallgrimsson, H., Jeong, H., Forde, D., Fineman, R., Raghuram, K.~J., Rehg, J.~M., and Ren, S.
\newblock Relcon: Relative contrastive learning for a motion foundation model for wearable data.
\newblock \emph{arXiv preprint arXiv:2411.18822}, 2024.

\bibitem[Xu et~al.(2018)Xu, Biswal, Deshpande, Maher, and Sun]{xu2018raim}
Xu, Y., Biswal, S., Deshpande, S.~R., Maher, K.~O., and Sun, J.
\newblock Raim: Recurrent attentive and intensive model of multimodal patient monitoring data.
\newblock In \emph{Proceedings of the 24th ACM SIGKDD International Conference on Knowledge Discovery \& Data Mining}, pp.\  2565--2573, 2018.

\bibitem[Yuan et~al.(2024)Yuan, Chan, Creagh, Tong, Acquah, Clifton, and Doherty]{yuan2024self}
Yuan, H., Chan, S., Creagh, A.~P., Tong, C., Acquah, A., Clifton, D.~A., and Doherty, A.
\newblock Self-supervised learning for human activity recognition using 700,000 person-days of wearable data.
\newblock \emph{npj Digital Medicine}, 7\penalty0 (1):\penalty0 91, 2024.

\bibitem[Zerveas et~al.(2021)Zerveas, Jayaraman, Patel, Bhamidipaty, and Eickhoff]{tsts2021}
Zerveas, G., Jayaraman, S., Patel, D., Bhamidipaty, A., and Eickhoff, C.
\newblock A transformer-based framework for multivariate time series representation learning.
\newblock In \emph{Proceedings of the 27th ACM SIGKDD Conference on Knowledge Discovery \& Data Mining}, pp.\  2114--2124, 2021.

\bibitem[Zhang \& Sennrich(2019)Zhang and Sennrich]{zhang2019root}
Zhang, B. and Sennrich, R.
\newblock Root mean square layer normalization.
\newblock \emph{Advances in Neural Information Processing Systems}, 32, 2019.

\bibitem[Zhang et~al.(2023)Zhang, Zheng, Cao, Bian, and Li]{zhang2023warpformer}
Zhang, J., Zheng, S., Cao, W., Bian, J., and Li, J.
\newblock Warpformer: A multi-scale modeling approach for irregular clinical time series.
\newblock In \emph{Proceedings of the 29th ACM SIGKDD Conference on Knowledge Discovery and Data Mining}, 2023.

\bibitem[Zheng et~al.(2023)Zheng, Chen, Sch{\"u}rch, Mollaysa, Allam, and Krauthammer]{zheng2023simts}
Zheng, X., Chen, X., Sch{\"u}rch, M., Mollaysa, A., Allam, A., and Krauthammer, M.
\newblock Simts: rethinking contrastive representation learning for time series forecasting.
\newblock \emph{arXiv preprint arXiv:2303.18205}, 2023.

\end{thebibliography}
\bibliographystyle{icml2025}

\appendix
\newpage
\onecolumn

\section{Appendix}

\subsection{Additional Dataset Details}

\subsubsection{Variable information}
\label{appendix:variable_info}

We list the different activity variables used as input modalities in our dataset in Table \ref{appendix:variable_list}. For certain quantities that can be captured by a watch or a phone, we separated these out into different variables. We also list how different variables are aggregated into hourly-level statistics when a quantity is measured multiple times within the hour. Variables that are cumulative (e.g., exercise time, step counts) are aggregated using a sum, while momentary variables (e.g., heart rate, respiratory rate) are aggregated using an average. The table also shows the frequency of collection of different variable types, both globally (i.e., what percentage of participant-weeks ever have an observed value for that variable), and per-subject (i.e., what percent of subjects ever have an observed value for that variable).
As expected, the most common variable types are quantities such as heart rate (observed in 99.9\% of subjects and 91.8\% of weeks) and activity metrics such as step count.
Less common variables were things like number of falls (only observed in less than 3\% of people, as most people do not experience a fall), or overnight wrist temperature (which is only collected overnight on more recent Apple Watch versions).
There is also substantial irregularity in the number of hourly variables recorded per week across the whole dataset. The median number of hourly variables recorded each week is 992, with an interquartile range of 829 to 1149, showing considerable variation in the number of observed variables each week. 

\begin{table}[!htb]
\centering
\small
\begin{tabular}{lllllll}
\toprule
\textbf{Variable} & \textbf{Category} & \textbf{\begin{tabular}[c]{@{}l@{}}Sampling \\ Rate\end{tabular}} & \textbf{\begin{tabular}[c]{@{}l@{}}Hourly\\ Aggregation\end{tabular}} & \textbf{Notes} & \textbf{\begin{tabular}[c]{@{}l@{}}\% Weeks\\ With 1+\\ Value\end{tabular}} & \textbf{\begin{tabular}[c]{@{}l@{}}\% Subjects\\ With 1+ \\ Value\end{tabular}} \\ \midrule
Flights climbed (phone) & Activity & \textless Hourly & Sum &  & 68.28 & 98.32 \\ 
Flights climbed (watch) & Activity & \textless Hourly & Sum &  & 65.06 & 97.59 \\ 
Active energy burned & Activity & \textless Hourly & Sum & \begin{tabular}[c]{@{}l@{}}Calories burned\\ while active\end{tabular} & 93.11 & 99.83 \\ 
Basal energy burned & Activity & \textless Hourly & Sum & \begin{tabular}[c]{@{}l@{}}Calories burned\\ at rest\end{tabular} & 95.21 & 99.50 \\ 
Step count (phone) & Activity & \textless Hourly & Sum &  & 98.34 & 99.33 \\ 
Step count (watch) & Activity & \textless Hourly & Sum &  & 90.67 & 99.49 \\ 
Exercise minutes & Activity & \textless Hourly & Sum &  & 86.35 & 99.64 \\ 
Stand time & Activity & \textless Hourly & Sum &  & 90.42 & 99.45 \\ 
Resting heart rate & Cardiovascular & Daily & Mean &  & 88.45 & 99.63 \\ 
Walking heart rate & Cardiovascular & Daily & Mean &  & 84.27 & 99.39 \\ 
Heart rate & Cardiovascular & \textless Hourly & Mean &  & 91.76 & 99.85 \\ 
Heart rate variability & Cardiovascular & \begin{tabular}[c]{@{}l@{}}Roughly \\ every few \\ hours\end{tabular} & Mean & \begin{tabular}[c]{@{}l@{}}Calculated via\\ Standard \\ Deviation \\ of Normal \\-to Normal \\ Interval \\ (SDNN)  \end{tabular} & 88.56 & 99.82 \\ 
Respiratory rate & Vitals & \textless Hourly & Mean & Overnight only & 35.21 & 60.79 \\ 
Blood oxygen saturation & Vitals & \begin{tabular}[c]{@{}l@{}}Roughly \\ every few\\ hours\end{tabular} & Mean & \begin{tabular}[c]{@{}l@{}}Only available\\ on Series 6+  \\ Apple Watch\end{tabular} & 44.31 & 62.56 \\ 
Wrist temperature & Vitals & Daily & Mean & \begin{tabular}[c]{@{}l@{}}Single value,\\ overnight only\end{tabular} & 3.12 & 10.42 \\ 
Walking speed & Mobility / Gait & \textless Hourly & Mean &  & 84.59 & 91.17 \\ 
Walking step length & Mobility / Gait & \textless Hourly & Mean &  & 84.53 & 91.09 \\ 
\begin{tabular}[c]{@{}l@{}}Walking double support\\ percentage\end{tabular} & Mobility / Gait & \textless Hourly & Mean &  & 83.23 & 90.86 \\ 
\begin{tabular}[c]{@{}l@{}}Walking asymmetry\\ percentage\end{tabular} & Mobility / Gait & \textless Hourly & Mean &  & 69.78 & 90.88 \\ 
Stair ascent speed & Mobility / Gait & \textless Hourly & Mean &  & 41.64 & 76.96 \\ 
Stair descent speed & Mobility / Gait & \textless Hourly & Mean &  & 41.72 & 76.79 \\ 
Fall count & Mobility / Gait & Opportunistic & Sum &  & 0.04 & 2.85 \\ 
Walking steadiness & Mobility / Gait & Weekly & Mean & \begin{tabular}[c]{@{}l@{}}An estimate of \\ stability while \\ walking\end{tabular} & 9.64 & 24.08 \\ 
Body mass & \begin{tabular}[c]{@{}l@{}}Body \\ Measurements\end{tabular} & Opportunistic & Mean & \begin{tabular}[c]{@{}l@{}}From third-party\\ devices or manual\\ input\end{tabular} & 10.93 & 54.76 \\ 
Body mass index & \begin{tabular}[c]{@{}l@{}}Body \\ Measurements\end{tabular} & Opportunistic & Mean & \begin{tabular}[c]{@{}l@{}}From third-party\\ devices or manual\\ input\end{tabular} & 7.97 & 33.84 \\ 
VO2max & \begin{tabular}[c]{@{}l@{}}Cardio Fitness / \\ Functional \\ Capacity\end{tabular} & Opportunistic & Mean & \begin{tabular}[c]{@{}l@{}}Requires outdoor\\ walk/run/hike\\ workouts\end{tabular} & 16.07 & 77.27 \\ 
6 minute walk distance & \begin{tabular}[c]{@{}l@{}}Cardio Fitness / \\ Functional \\ Capacity\end{tabular} & Weekly & Mean &  & 8.93 & 78.82 \\ \bottomrule
\end{tabular}
\caption{List of all derived health and behavioral quantity variables used in modeling, along with the method of aggregation used to create hourly-level aggregates, a brief description, the native sampling rate, and summary statistics (what percentage of individual weeks and subjects have at least one measurement).}
\label{appendix:variable_list}
\end{table}




\subsubsection{Information on participant self-reported medical history and medications surveys}
\label{appendix:survey}

We describe the different survey questions about medical questions in Table \ref{table: survey_medical_conditions} and medications in Table \ref{table: survey_medications}. These questions are found in the AHMS surveys and help define outcomes for participant-level health questions that we use to understand the information encoded in WBM. 

We also include two additional survey questions as downstream detection tasks: 1) Active Smoker (N = 23,8345): Did the participant answer "Yes" when asked about smoking status and "Every day" or "Some days" when asked about how often, or did they answer "No" when asked about smoking status, and "Not at all" when asked about how often, 2) On any medication (N = 23,943): Did the participant answer yes or no to the question "Do you currently take any form of medication"? 

\begin{table*}
      \centering
  \caption{AHMS survey questions about medical conditions. The main question is in form of `Have you ever been diagnosed with any of the following conditions?' and participants can answer `Yes' or `No' or `I prefer not to answer' or `I don't know'. The question for vision and hearing loss is different, which we explicitly mention in the corresponding rows. Third column indicates the number of left out participants for evaluation -- the reason for variations is that for each target we exclude participants whose answers were `I prefer not to answer' or `I don't know' or missing.}
  \small
  \begin{tabular}{p{3cm}p{7.5cm}p{1.2cm}p{1.2cm}} 
    \toprule
    \textbf{Target label} & \textbf{Medical condition} & \textbf{N (test)} & \textbf{\% Positive (test)}\\
    \midrule
    Heart attack & Heart attack (myocardial infarction)  & 27,009 & 1.56\\
    Heart disease & Coronary heart disease or angina pectoris & 26,765 & 2.45\\
    Blood pressure & High blood pressure (hypertension)& 26,519 & 25.11\\
    Stroke or TIA  & Stroke (cerebral hemorrhage, cerebral thrombosis) or transient ischemic attack (ministroke) & 27,022 & 1.45\\
    Afib & Atrial fibrillation & 26,457 & 3.22\\
    Heart rhythm & Heart rhythm problem other than atrial fibrillation & 26,264 & 8.86\\
    Pacemaker & Pacemaker & 27,128 & 0.56\\
    Artery disease & Peripheral artery disease & 26,627 & 0.74\\
    Heart failure & Heart failure & 27,051 & 0.95\\
     Diabetes & Diabetes & 26,861 & 6.67\\
    Cholesterol & High cholesterol & 26,414 & 27.00\\
    Arthritis & Arthritis & 26,588 & 18.38\\
    Hip/Knee & Hip or knee replacement & 27,153 & 2.54\\
    Lower back & Low back disorder or other chronic back defect & 26,646 & 16.27\\
    Neck disorder & Neck disorder or other chronic neck defect & 26,811 & 7.64\\
    Sleep apnea & Sleep apnea & 25,912 & 15.78\\
    Osteoporosis & Osteoporosis & 26,731 & 2.89\\
    Asthma & Asthma & 26,588 & 21.08\\
    Chronic bronchitis & Chronic bronchitis, chronic obstructive pulmonary disease, or emphysema & 26,919 & 3.57\\
    Allergy & Rhinitis, hay fever, eye inflammation, dermatitis, food allergy or other allergy (allergic asthma excluded) & 26,811 & 34.58\\
    Kidney & Kidney problems & 26,822 & 4.50\\
    Thyroid & Thyroid disease & 26,696 & 7.53\\
    Cancer & Cancer & 26,993 & 5.05\\
    Liver & Cirrhosis of the liver & 26,985 & 0.44\\
    Urinary & Urinary incontinence & 26,912 & 3.76\\
    Neuropathy & Neuropathy & 26,542 & 4.36\\
    Depression & Depression & 26,284 & 39.11\\
    Anxiety & Anxiety disorder & 26,167 & 37.97\\
    Hearing & Do you have hearing loss? & 24,457 & 15.64\\
    Vision & Do you have vision loss? & 25,663 & 25.15\\
    \bottomrule
  \end{tabular}
  \label{table: survey_medical_conditions}
\end{table*}

\begin{table*}
  \centering
  \caption{AHMS survey questions about medications. The main question is in form of `Do you currently take any of the following types of medications?' and participants can answer `Yes' or `No' or `I prefer not to answer'. The formatting for the medications is similar to their presentation in the tudy, but may not exactly match the format in the study application. Third column indicates the number of left out participants for evaluation -- the reason for variations is that for each target we exclude participants whose answers were `I prefer not to answer' or missing. Third party trademarks used herein are trademarks of their respective owners.}
  \small
  \begin{tabular}{p{3.3cm}p{7.3cm}p{1.2cm}p{1.2cm}} 
    \toprule
    \textbf{Target label} & \textbf{Medications} & \textbf{N (test)} & \textbf{\% Positive (test)}\\
    \midrule
    ACE-inhibitors & ACE-inhibitors or ARBs (for blood pressure) such as captopril, enalapril, lisinopril, losartan, ramipril, or valsartan & 15,736 & 18.44\\
    Anti-anxiety & Anti-anxiety aids such as alprazolam (Xanax\textsuperscript{\textregistered}), clonazepam (Klonopin\textsuperscript{\textregistered}), clorazepate (Tranxene\textsuperscript{\textregistered}), diazepam (Valium\textsuperscript{\textregistered}), or lorazepam (Ativan\textsuperscript{\textregistered}) & 15,742 & 17.50\\
    Anti-psychotics & Anti-psychotics such as haloperidol (Haldol\textsuperscript{\textregistered}), aripiprazole (Abilify\textsuperscript{\textregistered}), risperidone (Risperdal\textsuperscript{\textregistered}), quetiapine (Seroquel\textsuperscript{\textregistered}), olanzapine (Zyprexa\textsuperscript{\textregistered}), clozapine (Clozaril\textsuperscript{\textregistered}), or lurasidone (Latuda\textsuperscript{\textregistered}) & 15,780 & 3.97\\
    Anticoagulants  & Anticoagulants (blood thinners) such as warfarin (Coumadin\textsuperscript{\textregistered}), apixaban (Eliquis\textsuperscript{\textregistered}), betrixaban (Bevyxxa\textsuperscript{\textregistered}), dabigatran (Pradaxa\textsuperscript{\textregistered}), edoxaban (Lixiana\textsuperscript{\textregistered}), or rivaroxaban (Xarelto\textsuperscript{\textregistered}) & 15,774 & 3.77\\
    Antidepressants & Antidepressants such as amitriptyline (Elavil\textsuperscript{\textregistered}), bupropion (Wellbutrin\textsuperscript{\textregistered}), citalopram (Celexa\textsuperscript{\textregistered}), duloxetine (Cymbalta\textsuperscript{\textregistered}), escitalopram (Lexapro\textsuperscript{\textregistered}), fluoxetine (Prozac\textsuperscript{\textregistered}), paroxetine (Paxil\textsuperscript{\textregistered}), mirtazapine (Remeron\textsuperscript{\textregistered}), sertraline (Zoloft\textsuperscript{\textregistered}), or venlafaxine (Effexor\textsuperscript{\textregistered}) & 15,782 & 36.70\\
    Antiplatelets & Antiplatelets (blood thinners) such as aspirin, clopidogrel (Plavix\textsuperscript{\textregistered}), prasugrel (Effient\textsuperscript{\textregistered}), or ticagrelor (Brilinta\textsuperscript{\textregistered}) & 15,758 & 9.29\\
    Beta-blockers & Beta-blockers (for blood pressure or heart rhythm) such as atenolol (Tenormin\textsuperscript{\textregistered}), bisoprolol (Zebeta\textsuperscript{\textregistered}), carvedilol (Coreg\textsuperscript{\textregistered}), labetalol, metoprolol (Lopressor\textsuperscript{\textregistered}, Toprol-XL\textsuperscript{\textregistered}), nadolol (Corgard\textsuperscript{\textregistered}), nebivolol (Bystolic\textsuperscript{\textregistered}), propranolol (Inderal\textsuperscript{\textregistered}), or sotalol (Betapace\textsuperscript{\textregistered}) & 15,722 & 13.29\\
    Blood pressure med. & Other medications for lowering blood pressure such as clonidine, hydralazine, minoxidil, or sacubitril/valsartan (Entresto\textsuperscript{\textregistered}) & 15,694 & 4.82\\
    Calcium-channel blockers & Calcium-channel blockers (for blood pressure or heart rhythm) such as amlodipine (Norvasc\textsuperscript{\textregistered}), diltiazem, or verapamil & 15,677 & 7.13\\
     Chemotherapy & Certain types of chemotherapy such as carboplatin, cisplatin, oxaliplatin, vincristine, or vinblastine & 15,799 & 0.44\\
    Diuretics & Diuretics (water pills) such as chlorthalidone, furosemide (Lasix\textsuperscript{\textregistered}), hydrochlorothiazide, or spironolactone & 15,765 & 10.59\\
    Opioid painkillers & Opioid painkillers such as codeine, fentanyl, hydrocodone, hydromorphone (Dilaudid\textsuperscript{\textregistered}), meperidine (Demerol\textsuperscript{\textregistered}), morphine, oxycodone, Percocet\textsuperscript{\textregistered}, or Vicodin\textsuperscript{\textregistered} & 15,809 & 3.74\\
    Painkillers & Non-steroidal anti-inflammatories (painkillers) such as aspirin, celecoxib (Celebrex\textsuperscript{\textregistered}), diclofenac (Cambia\textsuperscript{\textregistered}), ibuprofen (Motrin\textsuperscript{\textregistered}/Advil\textsuperscript{\textregistered}), or naproxen (Aleve\textsuperscript{\textregistered}) & 15,788 & 44.14\\
    Sleep medication. & Sleeping aids such as eszopiclone (Lunesta\textsuperscript{\textregistered}), zaleplon (Sonata\textsuperscript{\textregistered}), or zolpidem (Ambien\textsuperscript{\textregistered}) & 15,753 & 9.35\\
    \bottomrule
  \end{tabular}
  \label{table: survey_medications}
\end{table*}

\subsubsection{Pre-Training Dataset Details}
\label{appendix:pretraining_dataset_info}

To create the full pre-training dataset, we filter to only weeks of data with at least 5 days of watch wear (determined by if there are heart rate samples on 5 unique days), and we preprocess such that each week begins at the same relative time, midnight on Monday. We further filter to include participants who have at least 5 usable weeks of data, and who have been enrolled in the study for at least 90 days.

We also show below in Table \ref{table:appendix_demog_details} the breakdown of demographics across all 161,855 subjects in the pre-training dataset. Given the large sample sizes, the demographics are nearly identical across the 80\% train, 10\% validation, and 10\% test splits so we report demographics across the full cohort in the table. Note that participants may self-report multiple race/ethnicity columns, so they do not sum to 100\%.

\begin{table}[]
\centering
\begin{tabular}{ll}
\toprule
\textbf{Covariate} & \textbf{Mean (IQR)} \\
\midrule
Age                & 40.7 (30.0 - 49.4)  \\
\midrule
Female             & 0.358               \\
Male               & 0.620               \\
Sex other or not set & 0.022             \\
\midrule
American Indian    & 0.022               \\
Asian              & 0.071               \\
Black              & 0.057               \\
Hispanic           & 0.112               \\
Middle Eastern     & 0.012               \\
White              & 0.791               \\
Ethnicity missing or preferred not to answer & 0.031 \\
Ethnicity none describe me & 0.013 \\
\midrule
BMI                & 28.6 (24.2 - 31.6)  \\    
\bottomrule
\end{tabular}
\caption{Demographics of the full cohort of 161,855 participants used for pretraining.}
\label{table:appendix_demog_details}   
\end{table}

\subsection{Additional Methods Details}
\label{appendix:methods}

\subsubsection{Setup}

 We assume one hour of wearable data at time $t$ from subject $s$ is a tuple of ($t^s, v^s_{t}, x^s_{t}$) where $t^s$ is the hour of recording, $v^s_{t}$ is the activity type or variable name, and $x^s_{t}$ is the value of the variable at time $t^s$. We denote a segment of wearable data as $d^s_{T} = \bigcup_{t \in T} (t^s, v^s_t, x^s_t)$, or the union of all observations within some week $T$. Irregularity from the signal comes from 1) events not being sampled at similar rates and 2) single events not being sampled consistently at the same rate. We z-score all input variables with respect to every observation in the training dataset and clip outliers. 

\subsection{Details on Tokenization of Wearable Data}
\label{appendix:tokenization}

In the main paper we consider two primary methods for tokenizing each week of data: 1) approaches that model each observation separately (\textbf{Tuple}), and 2) approaches that aggregate all information into a regularly sampled dense-matrix (\textbf{Dense}). 

\textbf{Tuple.} First, we consider mapping each tuple $(t^s, v^s_t, x^s_t) \in d^s_T$ to a $d$-dimension token $e^s_t$ that can be used as input to a sequential deep-learning model. To do so, we first consider learning mappings $\phi_x: x \longrightarrow \mathbb{R}^d$ and $\phi_v: v \longrightarrow \mathbb{R}^d$. From these vectors, we compute the $d$-dimensional input token $e^s_t$ as $\phi_v(v^s_t) + \phi_x(x^s_t)$ \citep{tuple2022,zhang2023warpformer,labach2023duett}. This results in a single input token for each unique behavioral measurement. A segment of wearable data can then be represented as a sequence of these learned $d$-dimensional tokens as $e^s_{T} = \bigcup_{t \in T} (e^s_t)$.  The tuple-based approach is well-suited for modeling irregular time-series, as it immediately handles irregular sampling for each variable type within a segment of wearable sensor data as well as missing values, as they are simply omitted. However, the tuple approach requires a separate token for each observation, which may not scale well for longer segments of wearable data, particularly when using memory-intensive modeling techniques. We further illustrate Tuple tokenization in Figure \ref{fig:tokenization}.

\textbf{Dense.} A potential limitation of tuple-based approaches is their poor scalablity and potential down-weighting of sparse but informative variables. Hence, we also consider the more traditional dense approach, where we convert data from tuples to a dense array of values. We create a matrix $X^s_T \in \mathcal{R}^{|T| \times |V|}$, where $|V|$ is the number of quantities or variables in the dataset and $|T|$ is the total number of possible hours per segment. For each tuple $(t^s, v^s_t, x^s_t) \in d^s_{T}$, we fill in the matrix $X^s_T$ with the value $x^s_t$ at index $(t^s, v^s_t)$. We also concatenate missingness indicators to the feature matrix, so that the final feature matrix is actually of size $|T| \times 2 \cdot |V| $. We next describe two different techniques for tokenizing the resulting matrix $X^s_T$ that can handle missingness.  

\textit{TST}. First, we consider the standard technique of imputing missingness values and treating the resulting matrix $X^s_T$ as a regularly sampled time-series. In initial experiments, we found that the simplest form of zero-imputation (i.e., global-average imputation with z-scored inputs) resulted in the strongest learned representations for a majority of meaning downstream tasks. We also tried other imputation approaches, such as using subject-specific means for each variable and week-specific means for each variable for a given subject (if available), but these did not work as well in practice.
Given this dense matrix representing a time series that is now regularly sampled, we learn a function that maps the vector of size $2 \cdot |V|$ from each hour of data to a $d$-dimensional vector $e^s_t \in \mathcal{R}^d$ using a multi-layer perceptron (MLP) as in past work (\textbf{TST}) \citep{tsts2021}. These input tokens $e^s_T = \{e^s_t\}_{t \in T}$ can then be used as input to an encoder for learning rich representations during pre-training. 

\textit{mTAN}. Second, we consider forgoing imputation and using multi-time attention with masking to tokenize the data (\textbf{mTAN}). We follow past work by \cite{mTAN2021} and \cite{primenet2023} and use learnable time-embeddings of size $d_T$ as queries $Q_T$ and keys $K_T$, and the embedding matrix $X^s_T$ as values through an attention computation. We then compute tokens via a masked attention computation as $e^s_{T} = (M^s_T \odot A^s_{T})X^s_T$, where $A^s_{T} = \text{softmax}(Q_{T}K_{T}/d_{T})$ and $M \in \{0,1\}^{|T| \times |V|}$ is a masking matrix where 1 means a value is observed, and 0 means a value is unobserved. The resulting columns of size $2 \cdot |V|$ can then be used as token inputs to a downstream deep learning encoder. 

\begin{figure}[!htb]
    \centering
    \includegraphics[width=1\columnwidth]{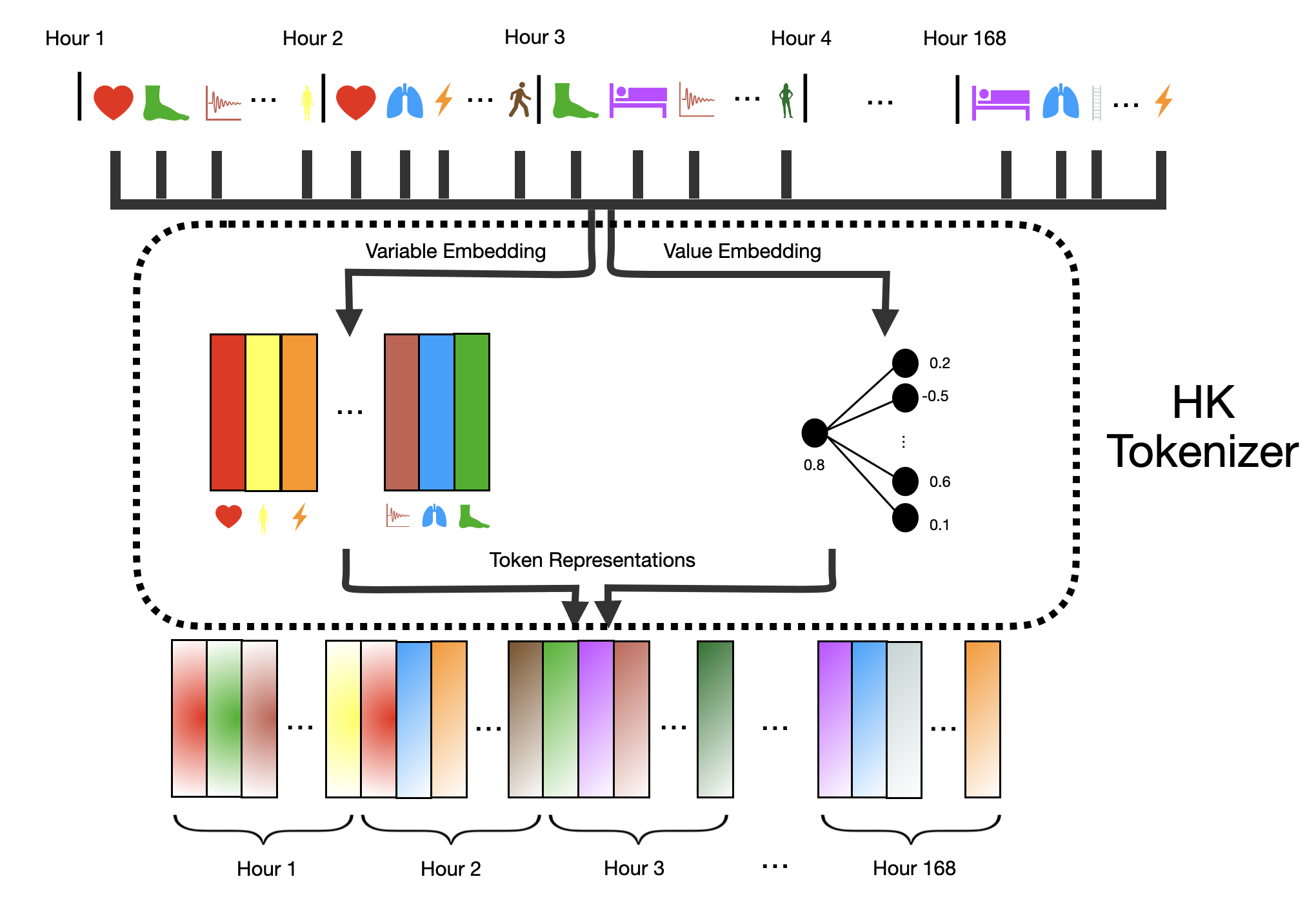}
    \caption{Tuple tokenization framework for irregularly sampled wearable activity data. Different colors refer to different variable types. Each variable is represented by a separate token unlike traditional multivariate modeling approaches. The token representation for each variable is the summation of its corresponding variable embedding and value embedding. For models that require learnable time embeddings, e.g. Transformer as in \cite{attention2017}, we also include time embeddings in the summation.}
    \label{fig:tokenization}
\end{figure}

\subsection{Details on Model Architectures}
\label{appendix:architectures}

Given the input segment $d^s_T$, the goal is to create a learned representation $r^s_T \in \mathbb{R}^d$ that summarizes information from the full segment of wearable activity data. 

\textbf{Self-Attention Transformer.} First, we consider standard self-attention techniques with learnable position encodings learned through some model $\phi_t: t \longrightarrow \mathbb{R}^d$. We augment each token $e^s_t$ with this learnable position encoding similar to past work through addition, where the new token is equal to $e^s_t + \phi_t(t)$ \citep{tuple2022}. 

\textbf{Rotary Transformer.} Next, we consider Rotary Position Embedding (RoPE), due to their strong performance compared to other variants, and their strong theoretical properties that show their ability to flexibly encode both absolute and relative encodings \citep{shaw2018self,ren2021rapt,dufter2022position,su2024roformer}. We modify the attention scheme of Transformers to apply RoPE embeddings at every layer using the hours / times $t$ as position IDs throughout.

For both schemes, we use a standard Transformer architecture, applying prior to attention and using a gated multi-layer perceptron and swish activation functions \citep{ramachandran2017swish,zhang2019root,liu2021pay}. To create the representation $r^s_T$ from the output of the Transformer, we calculate the average representation across all input tokens in the segment. 

\textbf{Mamba-2}. Finally, we consider using Mamba-2 as a backbone architecture to learn downstream representations. The Mamba-2 architecture takes in the same tokens as the Transformer without any additional positional encoding. As Mamba-2 is a recurrent model and the goal is to learn representations of the full signal, we apply a bi-directional Mamba-2 architecture as in past work using state-space models for time-series, allowing the representations to summarize information in both the forward and backward directions \citep{wang2024mamba,liang2024bi}. To create the single representation $r^s_T$, we first concatenate outputs of the forward and backward runs and use a single multi-layer perceptron to project this back to the original output space $\mathbb{R}^d$. From here, similar to the Transformer backbone, we calculate the average representation across all input tokens in the segment. 

\subsubsection{Regularized InfoNCE Loss}
\label{appendix:loss}

Given a batch of positive pairs of segments (i.e., pairs of segments from the same individual), we use both an InfoNCE loss to maximize similarity between positive segments and contrasting these segments from other examples in the batch, as well as a Kozachenko-Leonenko (KoLeo) regularization to help encourage a uniform span of features in a batch \citep{sablayrolles2018spreading,chen2020simple}. 

To calculate these losses, given a batch of $N$ positive pairs, we pass these through the tokenization and modeling architecture to obtain pairs of representations $(r^s_1, r^s_2)$. We map these representations to a new space of embeddings $(h^s_1, h^s_2)$ using a set of multi-layer perceptrons. We then calculate the one half of the InfoNCE loss as: 

$$L^{1,2}_{\text{InfoNCE}} = -\frac{1}{N} \sum_{i=1}^N \log \frac{\exp(sim(h^i_1, h^i_2)/\tau)}{\sum_{j=1}^N\exp(sim(h^i_1, h^j_2)/\tau)}\, ,$$ where $sim$ is the cosine similarity function. The InfoNCE loss encourages learning embeddings that are similar across positive pairs, and contrasts embeddings of segments from different subjects. To obtain symmetric results, we calculate the full InfoNCE loss as $$L_{\text{InfoNCE}}=\frac{1}{2}(L^{1,2}_{\text{InfoNCE}} + L^{2,1}_{\text{InfoNCE}})\, .$$

Next, to calculate one half of the KoLeo regularization, we compute: 

$$L^1_{\text{KoLeo}} = -\frac{1}{N} \sum_{i=1}^N \log (\min_{j \neq i} ||h^i_1 - h^j_1||^2)\, .$$ This loss encourages a uniform span of features across examples even further. For symmetry, we calculate the full regularization loss as $$L_{\text{KoLeo}} = \frac{1}{2}(L^1_{\text{KoLeo}} + L^2_{\text{KoLeo}})\, .$$

The final self-supervised pretraining loss over each batch can finally be computed as $$L_{\text{InfoNCE}} + \lambda L_{\text{KoLeo}},$$ where $\lambda$ is a hyperparameter that controls the trade-off between the contrastive and regularization terms. 

\subsection{Additional Experimental Setup Details}

\subsubsection{Modeling and Hyperparameter Choices}
\label{appendix:hyps}

For all models, we use the AdamW optimization with a learning rate of 0.001 and a sequential learning rate schedule. The learning rate begins with a linear warm up with a start factor of 0.5, followed by an exponential strategy with a gamma of 0.995. All backbones use a feed-forward hidden dimension size (H) of 4 times the input hidden dimension (D) for feed-forward layers inside of the model block, and all Transformer models use 8 attention heads. To map the final representation from each backbone to a new dimension for loss calculation, we use a feed-forward neural network of three layers of size input hidden dimension D, input hidden dimension D$\times$ 4, and input hidden dimension D, with batch normalization and a dropout of 30\%.

We conduct two separate hyperparameter search jobs. First, we perform a smaller hyperparameter search across all 9 combinations of models. We describe the different hyperparameters in Table \ref{table:hyp_ablation}. For the ablation test, we use an input hidden dimension D of size 128. 

For the final WBM model, we perform a larger hyperparameter search over the space defined in Table \ref{table:hyp_big}. For the final model, we use a input hidden dimension D of size 256. For all models with a Mamba-2 backbone, the number of layers denotes the full bi-directional model. 

The final WBM model is a Mamba-2 backbone with TST tokenization. We use a batch size of 192 samples, hidden dimension of size 256, a feed-forward hidden dimension of 1024, weight decay of 0.035, 24 backbone layers, dropout of 2.7\%, layer normalization instead of RMS normalization, a $\lambda$ of 0.21 for KoLeo regularization, and dropping 23.3\% of input tokens to create augmentations during training. The model was trained with the AdamW optimizer using the settings discussed above. The final WBM model was the result of 6 epochs of training which took 16 hours of training time on 8 A100 GPUs. The learned model can quickly perform inference, and embeddings can be used easily across many tasks.

\begin{table}[!htb]
\centering
\begin{tabular}{cc}
\toprule
\textbf{Hyperparameter}       & \textbf{Search Space} \\ \midrule
Batch Size    & \{128, 192\}  \\ 
Number of Layers                   & \{12, 16\}\\ 
Weight Decay & \{0.01, 0.001\}    \\ 
$\lambda$ (KoLeo Regularization)   & \{0.1 ,0.2\}  \\ \bottomrule
\end{tabular}
\caption{Hyperparameter search for ablation experiments.}
\label{table:hyp_ablation}
\end{table}

\begin{table}[!htb]
\centering
\begin{tabular}{cc}
\toprule
\textbf{Hyperparameter}       & \textbf{Search Space} \\ \midrule
Batch Size    & \{128, 192, 256\}  \\ 
Number of Layers                   & \{8, 12, 16, 20, 24\} \\ 
Weight Decay & [0.0001, 0.1]    \\ 
$\lambda$ (KoLeo Regularization)   & [0.05, 0.4]  \\ 
Dropout   & [0\%, 70\%]  \\ 
Token Drop Percentage   & [10\%, 70\%]  \\ 
Normalization  & \{RMSNorm, LayerNorm\} \\ \bottomrule
\end{tabular}
\caption{Hyperparameter search for final WBM model training.}
\label{table:hyp_big}
\end{table}

\subsubsection{Ablation Results}
\label{ablation}

First, we present results on a validation set for the downstream task of age prediction. We report mean absolute error (MAE) for predicting age in Table \ref{table:hk_ablation}. Most models perform similarly, with the Tuple tokenization with Mamba-2 performing the worst. However, the TST tokenization with a Mamba-2 backbone resulted in the best performance overall for this task. 

\begin{table}[!htb]
\centering
\begin{tabular}{cc}
\toprule
\textbf{Embedding}       & \textbf{Age MAE $\downarrow$} \\ \midrule
Tuple with Self-Attention     & 4.65   \\ 
Tuple with Rotary Attention                    & 4.29 \\ 
Tuple with Mamba-2 & 6.21     \\ 
mTAN with Self-Attention   & 4.51  \\ 
mTAN with Rotary Attention   & 4.39  \\ 
mTAN with Mamba-2   & 4.73  \\ 
TST with Self-Attention  & 4.37  \\ 
TST with Rotary Attention   & 4.43 \\ 
\textbf{TST with Mamba-2}   & \textbf{4.05}  \\ \bottomrule
\end{tabular}
\caption{Mean absolute error for age prediction for held-out validation task across different foundation model architectures in an initial ablation study.}
\label{table:hk_ablation}
\end{table}

We present further results on the ablation of different tokenization and deep learning architectures. In Table \ref{table:hk_ablation_survey}, we show the number of times each combination of tokenization and deep learning model was the best performing mode for the 47 survey outcomes presented in the main paper. The relative ordering of model performance holds, with the TST with Mamba-2 model performing best. These results provide more evidence that the TST with Mamba-2 framework is the strongest paradigm for training the final foundation model. 

\begin{table*}[!htb]
\centering
\begin{tabular}{cc}
\toprule
\textbf{Embedding}       & \textbf{Number of Outcomes with Best Performance} \\ \midrule
Tuple with Self-Attention     & 1/47   \\ 
Tuple with Rotary Attention                    & 6/47 \\ 
Tuple with Mamba-2 & 0/47    \\ 
mTAN with Self-Attention   & 1/47  \\ 
mTAN with Rotary Attention   & 2/47  \\ 
mTAN with Mamba-2   & 0/47  \\ 
TST with Self-Attention  & 10/47  \\ 
TST with Rotary Attention   & 2/47\\ 
\textbf{TST with Mamba-2}   & 25/47  \\ \bottomrule
\end{tabular}
\caption{The number of survey outcomes for which each embedding was the best performing model. The relative performance of models is consistent with the age prediction performance, with TST with Mamba-2 performing best.}
\label{table:hk_ablation_survey}
\end{table*}

\subsubsection{Masked Autoencoder Results}
\label{appendix:MAE}

Due to the popularity of masked autoencoder approaches, we consider training the strongest backbone model (i.e., TST tokenization with Mamba-2) using a masked autoencoder approach. We use Mamba-2 as both the encoder and decoder, and perform a similar hyperparameter search over the variables mentioned in Table \ref{table:hyp_big}, along with searching over the percentage of signal to mask. We ignore missing variables in the calculation of the loss and try both random masking and temporal masking, akin to past work \citep{narayanswamy2024scaling}. The best model across all settings achieves an MAE of 6.39 on the task of age prediction, which is substantially worse than all model strained using the contrastive loss. We hypothesize that this could be due to the restrictive nature of masked autoencoder approaches in requiring the entire input signal to be reconstructed. This could result in embeddings that focus too much on reconstructing the most observed variables (e.g., active energy burned, step count, etc), rather than learning embeddings that learn a holistic view of an individual's week of data, including sparse yet informative variables (e.g. VO2max, six minute walk distance, etc). Techniques such as joint embedding predictive architectures (JEPA) \citep{zheng2023simts,assran2023self} could be an interesting choice to overcome this problem; we leave this exploration for future work. 

\subsection{Additional Downstream Task Details}

\subsubsection{Task Preprocessing Details}
\label{appendix:task_details}

\paragraph{Demographics:} The first downstream task is to predict participant age and biological sex (male or female).
Prior work has shown that biosignals can often reliably predict demographics \citep{apple_ppg_FM_2024}, and since demographics are fairly predictive of many health conditions, this offers a simple proxy task for assessing model quality.
All 161,855 participants have a self-reported age, and 158,347 have a self-reported biological sex of either male or female (63.4\% male). We use the age prediction task to find the best combination of settings in the initial ablation analysis. 

\paragraph{Baseline disease and medications:}
As in \citet{apple_ppg_FM_2024}, we use the intake surveys that ask questions about participant baseline disease and medication history as downstream tasks.
In order to be included in the dataset for each task, a participant needs to have responded with either yes or no as to whether or not they had the disease or were taking the medication type upon enrollment; participants declining to respond or answering unsure are removed for that task.
In total, we consider 47 survey questions as outcomes. The full set of exact survey questions and counts for each condition can be found in Appendix \ref{appendix:survey}, while results figures show a shortened version of the name for each prediction task.

\paragraph{Sleep metrics:}
Apple Watch can be used to estimate stages of sleep for users who wear a watch overnight, and are derived from an accelerometer-based algorithm \cite{apple2023sleep}. 
Since our foundation models were not trained using any sleep data, we use weekly summaries of sleep data (in particular, the weekly average total sleep duration, deep sleep duration, REM sleep duration, and sleep efficiency -- the percent of time not in a wake state) as sleep-related downstream tasks, to assess the ability of the models to generalize to a new setting.
We use data from 35.2K unique users who have sleep tracking data in AHMS.
There are 671.6K unique participant-weeks of sleep outcomes data that we use, after limiting to weeks where at least 5 nights of sleep were recorded, and where there was sufficient HealthKit data as well as at least one PPG segment.

\paragraph{Pregnancy:}
It is well known that many health metrics change during pregnancy, some of which are directly measurable by wearables, such as resting heart rate and HRV \citep{mishra2024trends}.
Hence, prediction of whether or not a week of data is from during a period when a participant is pregnant is a health state that should be detectable by our models.
To create a pregnancy dataset, we limit to the 430 pregnancies ending in a vaginal or cesarean delivery from 385 unique participants, among participants with adequate HealthKit and PPG data.
Weeks in the 9 months before the delivery date and 1 month after are categorized as ``positive'', in the sense that these weeks may show physiological changes for these participants, and all other periods of time are ``negative''.
We further include 24,225 female participants younger than 50 who have reasonable coverage in their health surveys as non-pregnant controls.

\paragraph{Infection:}
Since substantial prior work has shown that wearables can detect influenza-like illness and COVID-19 \citep{nestor2023machine}, we we also use this as a downstream task.
We predict whether a week of time series data is from within 1 week before to 4 weeks after self-reporting a respiratory infection, such as the common cold, flu, or COVID-19.
Other weeks not in proximity to a reported respiratory infection are treated as negatives.
We only include weeks of data from participants who reliably fill out the quarterly/monthly health questionnaires.
The task dataset includes 9689 self-reported new respiratory conditions (e.g. cold, COVID-19, flu, pneumonia) that developed in 8585 unique participants.
We also create additional negative controls from participants who reliably fill out surveys but never self-report a respiratory infection, yielding an additional 83K participants all labeled as negative.

\paragraph{A1C / Diabetes Status:}
To assess how well our models encode information about metabolic health, we predict diabetes status using HbA1c, a laboratory value typically used for diabetes diagnosis.
These values are ascertained from clinical health records, so should be higher quality than the self-reported labels of having diabetes on enrollment into the study.
There are 17,984 unique HbA1c labs from 9,847 unique participants, with 116.5K weeks of data collected within 30 days of each lab value that we use to make predictions.
The prediction task is to classify a participant as either having diabetes or not, based upon the standard threshold of $\geq 6.5\%$ for the HbA1c value.
In our dataset, 53.2\% of labs were normal $(< 5.7\%)$, 26.5\% were prediabetes ($\geq 5.7\%$ and $<6.5\%)$, and 20.3\% were diabetes ($\geq 6.5\%)$.

\paragraph{Injury:}
A task that we would expect WBM to outperform PPG on is predicting metrics related to changes in mobility, such as whether a participant recently self-reported an accident or injury limiting their mobility.
We would expect such events to affect mobility metrics captured in activity data, but may not be manifested in a biosignal like PPG.
We create a dataset of 26K self-reported injuries that happened during the study to 17.7K unique individuals. 
We label the week of the injury and the 4 weeks after as positive weeks, omit weeks 5-16 after injury (in the event that the person recovers quickly, to avoid the potential for ambiguous labels), and otherwise label all other weeks from these participants as negatives.
We also include as negative controls data from 75.7K other participants who have adequate data availability and survey coverage but never self-report an injury, with all of their data labeled as negative.

\subsubsection{Additional Results}
\label{appendix:additional_results}

\textbf{Reconstructing Input Signals}. To further understand the information encoded in WBM, we probe the embeddings to predict the weekly mean of each input activity variable (Table \ref{table:hk_stats_embed_preds}). Specifically, we are interested in understanding whether the embeddings can only reconstruct the highly observed input variables. However, the results show that the embeddings are predictive of a majority of input variables, including irregularly sampled variables such as stair ascent and descent speeds, walking heart rates, and VO2 max. Meanwhile, active energy burned, a frequently measured variable, is unable to be constructed using the embeddings, providing evidence against the hypothesis that the embeddings simply capture the most frequently observed variables. This is likely an artifact of our contrastive pre-training objective -- a different pre-training paradigm such as masked autoencoding would likely be better able to reconstruct such a frequently observed variable. We hypothesize that since active energy burned tends to correlate highly with heart rate, the contrastively trained model simply learns to discard it. Moreover, we hypothesize that we may be able to remove some of these features that do not contribute to the final representation during the training process to get smaller inputs and gain computational efficiency.

\begin{table}[!htb]
\centering
\begin{tabular}{cc}
\toprule
\textbf{Activity Metric}       & $R^2$ Performance  \\ \midrule
Resting heart rate, mean & 0.942 \\ 
Heart rate, mean     & 0.938  \\ 
VO2max, mean & 0.929  \\ 
Walking double support percentage, mean & 0.909 \\ 
Body mass, mean & 0.900 \\ 
Walking HR, mean & 0.888 \\ 
Stand time, mean & 0.872 \\ 
HRV (SDNN), mean & 0.868 \\ 
Walking step length, mean & 0.866  \\ 
Walking speed, mean & 0.849  \\ 
Step count (watch), mean & 0.847 \\ 
Respiratory rate, mean & 0.840  \\ 
Walking Steadiness, mean & 0.810 \\ 
Six minute walk test distance, mean & 0.796 \\ 
Step count (phone), mean & 0.793  \\ 
Stair descent speed, mean & 0.752  \\ 
Exercise minutes, mean & 0.746 \\ 
Sleeping wrist temp, mean & 0.612  \\ 
Stair ascent speed, mean & 0.563\\ 
Walking asymmetry percentage, mean & 0.533  \\ 
Flights climbed (watch), mean & 0.418 \\ 
Basal energy burned, mean & 0.347 \\ 
Flights climbed (phone), mean & 0.341 \\ 
Number of times fallen, mean & 0.096 \\ 
Active energy burned, mean & 0.011 \\ 
Body mass index, mean & 0.000 \\ 
Oxygen saturation, mean & 0.000 \\ \bottomrule
\end{tabular}
\caption{$R^2$ performance of WBM embeddings for predicting the weekly mean of all activity variables. We subset this analysis to only weeks where a given HK variable is actually observed, and so has a mean value, in order to ignore artifacts from imputed values skewing the results. Our results show that even irregularly sampled activity variables are captured within the embeddings.}
\label{table:hk_stats_embed_preds}
\end{table}

\newpage
\textbf{Results Across Demographic Subgroups.} We report results of trained models across demographic subgroups on representative tasks both at a subject-level and a segment-level. We report performance across different race/ethnicity categories, age subgroups, and biological sex. For race/ethnicity categories, we note that many subjects may fall into multiple race/ethnicity categories as subjects can self-report belonging to more than one. These subjects will be considered in multiple evaluations. 

For subject-level tasks, we report performance for heart failure, calcium-channel blockers, and smoking status. Results show that most models perform well across different demographic subgroups besides the baseline model. However, this is to be expected as the baseline uses this demographic information as input to the model. Hence, stratifying on these groups results in poor discriminative performance. One exception to the strong performance across subgroups is the WBM model on the heart failure outcome. Both models using the WBM embeddings (WBM and WBM + PPG) see poor performance compared to the PPG model, likely indicating that these demographics are encoded and being used in the final learned model. 

For intra-subject tasks, we report performance for predicting respiratory infections and pregnancy. Again, we see that most models generally perform well across demographic subgroups except for the baseline model. One notable exception is on predicting pregnancy within Black patients, where the baseline model excels.

\begin{table}[!htb]
\small
\centering
\begin{tabular}{lllll}
\toprule
Age Group & Baseline & WBM & PPG & WBM + PPG \\
\midrule
$<=$ 33 years ( N ( \%): 8922 (0.33)) & 0.560 (0.468, 0.643) & 0.618 (0.519, 0.715) & \textbf{0.751 (0.650, 0.840) }& 0.737 (0.639, 0.830) \\
34 - 46 years ( N (\%): 9264 (0.63)) & 0.569 (0.508, 0.634) & 0.763 (0.700, 0.821) & \textbf{0.855 (0.797, 0.908)} & 0.842 (0.781, 0.897) \\
$>$ 47 years( N (\%): 8865 (1.91)) & 0.593 (0.554, 0.634) & 0.815 (0.786, 0.845) & \textbf{0.889 (0.862, 0.914)} & 0.884 (0.857, 0.911) \\
\bottomrule
\end{tabular}
\caption{AUROC performance on heart failure across different age subgroups.}
\label{table:hf_age}
\end{table}

\begin{table}[!htb]
\small
\centering
\begin{tabular}{lllll}
\toprule
Ethnicity & Baseline & WBM & PPG & WBM + PPG \\
\midrule
American Indian (N (\%): 599 (1.50)) & 0.592 (0.424, 0.756) & \textbf{0.799 (0.631, 0.943)} & 0.760 (0.555, 0.941) & 0.780 (0.586, 0.941) \\
Asian (N (\%): 1702 (0.41)) & 0.450 (0.201, 0.688) & 0.768 (0.540, 0.988) & \textbf{0.857 (0.721, 0.986)} & 0.828 (0.653, 0.982) \\
Black (N (\%): 1483 (1.21)) & 0.562 (0.438, 0.679) & 0.773 (0.680, 0.868) & \textbf{0.818 (0.696, 0.920)} & 0.811 (0.683, 0.919) \\
Hispanic (N (\%): 2886 (0.80)) & 0.611 (0.511, 0.710) & 0.734 (0.589, 0.858) & \textbf{0.903 (0.822, 0.967)} & 0.900 (0.812, 0.966) \\
Middle Eastern (N (\%): 315 (1.27)) & 0.235 (0.067, 0.387) & 0.775 (0.559, 0.958) & \textbf{0.797 (0.560, 0.920)} & 0.727 (0.405, 0.906) \\
White (N (\%): 21749 (1.02)) & 0.624 (0.592, 0.655) & 0.812 (0.784, 0.842) & \textbf{0.879 (0.849, 0.906)} & 0.873 (0.842, 0.900) \\
\bottomrule
\end{tabular}
\caption{AUROC performance on heart failure across different race/ethnicity subgroups.}
\label{table:hf_ethnicity}
\end{table}

\begin{table}[!htb]
\small
\centering
\begin{tabular}{lllll}
\toprule
Biological Sex & Baseline & WBM & PPG & WBM + PPG \\
\midrule
Male (N (\%): 16370 (0.95)) & 0.632 (0.590, 0.672) & 0.830 (0.790, 0.864) & \textbf{0.885 (0.849, 0.918)} & 0.884 (0.849, 0.917) \\
Female (N (\%): 10102 (0.92)) & 0.597 (0.544, 0.647) & 0.802 (0.758, 0.842) & \textbf{0.881 (0.846, 0.912)} & 0.879 (0.843, 0.910) \\
\bottomrule
\end{tabular}

\caption{AUROC performance on heart failure across different biological sex subgroups.}
\label{table:hf_sex}
\end{table}

\begin{table}[!htb]
\small
\centering
\begin{tabular}{lllll}
\toprule
Age Group & Baseline & WBM & PPG & WBM + PPG \\
\midrule
$<= 35$ years ( N (\%): 5082 (1.75)) & 0.460 (0.400, 0.524) & \textbf{0.682 (0.624, 0.742)} & 0.659 (0.599, 0.717) & 0.656 (0.596, 0.719) \\
36 - 49 years ( N (\%): 5307 (5.88)) & 0.492 (0.460, 0.525) & \textbf{0.749 (0.720, 0.776)} & 0.732 (0.702, 0.761) & 0.735 (0.706, 0.765) \\
$> 50$ years ( N (\%): 5288 (13.54)) & 0.485 (0.463, 0.507) & 0.715 (0.695, 0.734) & 0.727 (0.707, 0.747) & \textbf{0.733 (0.714, 0.752}) \\
\bottomrule
\end{tabular}
\caption{AUROC performance on calcium-channel blockers across different age subgroups.}
\label{table:ccb_age}
\end{table}

\begin{table}[!htb]
\small
\centering
\begin{tabular}{lllll}
\toprule
Ethnicity & Baseline & WBM & PPG & WBM + PPG \\
\midrule
American Indian (N (\%): 348 (10.34)) & 0.614 (0.523, 0.701) & 0.825 (0.753, 0.884) & \textbf{0.837 (0.762, 0.903)} & 0.827 (0.750, 0.895) \\
Asian (N (\%): 712 (7.16)) & 0.513 (0.435, 0.597) & 0.675 (0.598, 0.748) & 0.775 (0.709, 0.840) & \textbf{0.784 (0.722, 0.847}) \\
Black (N (\%): 754 (13.40) & 0.612 (0.558, 0.671) & 0.763 (0.708, 0.815) & 0.788 (0.740, 0.832) & \textbf{0.794 (0.745, 0.840)} \\
Hispanic (N (\%): 1357 (5.90)) & 0.494 (0.433, 0.548) & 0.786 (0.731, 0.837) & \textbf{0.809 (0.753, 0.860)} & 0.786 (0.727, 0.846) \\
Middle Eastern (N (\%): 163 (6.13)) & 0.597 (0.467, 0.718) & 0.731 (0.509, 0.922) & \textbf{0.780 (0.574, 0.948)} & 0.765 (0.541, 0.962) \\
White (N (\%): 13262 (6.83)) & 0.523 (0.504, 0.541) & \textbf{0.796 (0.780, 0.810)} & 0.782 (0.765, 0.800) & 0.778 (0.760, 0.796) \\
\bottomrule
\end{tabular}

\caption{AUROC performance on calcium-channel blockers across different race/ethnicity subgroups.}
\label{table:ccb_ethnicity}
\end{table}

\begin{table}[!htb]
\small
\centering
\begin{tabular}{lllll}
\toprule
Biological Sex & Baseline & WBM & PPG & WBM + PPG \\
\midrule
Male (N (\%): 8607 (9.00)) & 0.521 (0.502, 0.542) & 0.770 (0.753, 0.788) & 0.786 (0.769, 0.804) & \textbf{0.791 (0.774, 0.809)} \\
Female (N (\%): 6717 (4.69)) & 0.510 (0.481, 0.542) & \textbf{0.796 (0.771, 0.823)} & 0.791 (0.762, 0.820) & 0.789 (0.761, 0.817) \\
\bottomrule
\end{tabular}
\caption{AUROC performance on calcium-channel blockers across different biological sex subgroups.}
\label{table:ccb_sex}
\end{table}

\begin{table}[!htb]
\small
\centering
\begin{tabular}{lllll}
\toprule
Age Group & Baseline & WBM & PPG & WBM + PPG \\
\midrule
$<=$ 34 years ( N (\%): 8253 (7.22)) & 0.636 (0.614, 0.658) & \textbf{0.803 (0.785, 0.821)} & 0.789 (0.770, 0.811) & 0.795 (0.776, 0.816) \\
35.0-46.0 ( N (\%): 7473 (8.71)) & 0.654 (0.632, 0.675) & \textbf{0.824 (0.806, 0.840)} & 0.809 (0.788, 0.828) & 0.816 (0.794, 0.834) \\
$>$ 47 years ( N (\%): 8109 (4.72)) & 0.695 (0.669, 0.719) & 0.793 (0.767, 0.818) & 0.808 (0.782, 0.834) & \textbf{0.813 (0.787, 0.838)} \\
\bottomrule
\end{tabular}
\caption{AUROC performance on smoker status across different age subgroups.}
\label{table:smoke_age}
\end{table}

\begin{table}[!htb]
\small
\centering
\begin{tabular}{lllll}
\toprule
Ethnicity & Baseline & WBM & PPG & WBM + PPG \\
\midrule
American Indian (N (\%): 511 (13.31)) & 0.623 (0.560, 0.688) & 0.819 (0.771, 0.864) & 0.834 (0.786, 0.883) & \textbf{0.846 (0.799, 0.892)} \\
Asian (N (\%): 1425 (6.95)) & 0.649 (0.593, 0.705) & \textbf{0.809 (0.762, 0.850)} & 0.717 (0.659, 0.779) & 0.732 (0.675, 0.791) \\
Black (N (\%): 1248 (8.41)) & 0.536 (0.479, 0.597) & 0.793 (0.749, 0.837) & 0.807 (0.759, 0.850) & \textbf{0.823 (0.776, 0.865)} \\
Hispanic (N (\%): 2431 (6.83)) & 0.614 (0.568, 0.661) & \textbf{0.780 (0.741, 0.819)} & 0.740 (0.698, 0.785) & 0.759 (0.718, 0.804) \\
Middle Eastern (N (\%): 279 (11.11)) & 0.666 (0.561, 0.774) & 0.773 (0.691, 0.843) & 0.827 (0.756, 0.896) & \textbf{0.838 (0.771, 0.901)} \\
White (N (\%): 19385 (6.61)) & 0.674 (0.659, 0.688) & 0.816 (0.804, 0.828) & 0.804 (0.790, 0.817) & \textbf{0.822 (0.809, 0.834)} \\
\bottomrule
\end{tabular}

\caption{AUROC performance on smoker status across different race/ethnicity subgroups.}
\label{table:smoke_ethnicity}
\end{table}

\begin{table}[!htb]
\small
\centering
\begin{tabular}{lllll}
\toprule
Biological Sex & Baseline & WBM & PPG & WBM + PPG \\
\midrule
Male (N (\%): 14464 (6.34)) & 0.632 (0.614, 0.651) & \textbf{0.802 (0.786, 0.817)} & 0.774 (0.757, 0.791) & 0.783 (0.766, 0.800) \\
Female (N (\%)): 8863 (7.71) & 0.699 (0.681, 0.719) & 0.841 (0.826, 0.857) & 0.827 (0.809, 0.844) & \textbf{0.856 (0.840, 0.872)} \\
\bottomrule
\end{tabular}

\caption{AUROC performance on smoker status across different biological sex subgroups.}
\label{table:smoke_sex}
\end{table}

\begin{table}[!htb]
\small
\centering
\begin{tabular}{lllll}
\toprule
Ethnicity & Baseline & WBM & PPG & WBM + PPG \\
\midrule
American Indian (N (\%): 14,359 (1.1)) & 0.678 (0.637, 0.712) & \textbf{0.717 (0.678, 0.754)} & 0.692 (0.657, 0.733) & 0.702 (0.660, 0.738) \\
Asian (N (\%):  39,501 (0.7)) & 0.567 (0.533, 0.604) & \textbf{0.766 (0.743, 0.790)} & 0.721 (0.693, 0.750) & 0.756 (0.731, 0.783) \\
Black (N (\%):  31,822 (0.9)) & 0.629 (0.603, 0.657) & \textbf{0.645 (0.616, 0.675)} & 0.637 (0.607, 0.675) & 0.633 (0.600, 0.664) \\
Hispanic (N (\%):  67,190 (0.9)) & 0.572 (0.552, 0.597) & 0.677 (0.658, 0.696) & \textbf{0.697 (0.677, 0.718)} & 0.690 (0.674, 0.712) \\
Middle Eastern (N (\%):  7,550 (0.6)) & 0.663 (0.574, 0.754) & 0.586 (0.527, 0.655) & \textbf{0.673 (0.607, 0.738)} & 0.652 (0.590, 0.739) \\
White (N (\%):  666,408 (0.9)) & 0.628 (0.621, 0.634) & 0.749 (0.743, 0.756) & 0.729 (0.723, 0.735) & \textbf{0.762 (0.756, 0.768)} \\
\bottomrule
\end{tabular}
\caption{Test-set AUROC performance for predicting respiratory infections across different race/ethnicity subgroups. The sample sizes presented are per-week.}
\label{table:inf_ethnicity}
\end{table}

\begin{table}[!htb]
\small
\centering
\begin{tabular}{lllll}
\toprule
Biological Sex & Baseline & WBM & PPG & WBM + PPG \\
\midrule
Male (N (\%): 491,031 (0.8)) & 0.627 (0.619, 0.635) & 0.759 (0.753, 0.766) & 0.745 (0.739, 0.752) & \textbf{0.776 (0.770, 0.783)} \\
Female (N (\%): 278,711 (1.1)) & 0.630 (0.621, 0.638) & 0.725 (0.716, 0.733) & 0.692 (0.684, 0.699) & \textbf{0.727 (0.719, 0.735)} \\
\bottomrule
\end{tabular}
\caption{Test-set AUROC performance for predicting respiratory infections across different biological sex subgroups. The sample sizes presented are per-week.}
\label{table:inf_sex}
\end{table}

\begin{table}[!htb]
\small
\centering
\begin{tabular}{lllll}
\toprule
Age & Baseline & WBM & PPG & WBM + PPG \\
\midrule
$<=$ 37 years (N (\%): 269,084 (0.8)) & 0.615 (0.604, 0.628) & 0.738 (0.727, 0.747) & 0.707 (0.698, 0.718) & \textbf{0.741 (0.731, 0.751)} \\
38 - 51 years (N (\%): 262,574 (1.0)) & 0.641 (0.629, 0.651) & 0.752 (0.742, 0.760) & 0.728 (0.719, 0.737) & \textbf{0.757 (0.748, 0.765)} \\
$>$ 52 years (N (\%): 255,629 (1.0)) & 0.633 (0.622, 0.643) & 0.749 (0.740, 0.759) & 0.724 (0.716, 0.734) & \textbf{0.760 (0.752, 0.771)}\\
\bottomrule
\end{tabular}
\caption{Test-set AUROC performance for predicting respiratory infections across different age subgroups. The sample sizes presented are per-week.}
\label{table:inf_age}
\end{table}

\begin{table}[!htb]
\small
\centering
\begin{tabular}{lllll}
\toprule
Ethnicity & Baseline & WBM & PPG & WBM + PPG \\
\midrule
Asian (N (\%):  7,023 (1.7)) & 0.582 (0.543, 0.622) & 0.566 (0.506, 0.628) & \textbf{0.763 (0.721, 0.800)} & 0.737 (0.685, 0.786) \\
Black (N (\%):  9,897 (0.5)) & \textbf{0.823 (0.785, 0.852)} & 0.764 (0.690, 0.819) & 0.532 (0.459, 0.603) & 0.713 (0.643, 0.776) \\
Hispanic (N (\%):  17,718 (1.2)) & 0.687 (0.651, 0.724) & 0.716 (0.683, 0.753) & 0.790 (0.757, 0.823) & \textbf{0.841 (0.807, 0.870)} \\
White (N (\%):  153,470 (1.1)) & 0.817 (0.808, 0.828) & 0.879 (0.870, 0.887) & 0.879 (0.870, 0.888) & \textbf{0.926 (0.919, 0.933)} \\
\bottomrule
\end{tabular}
\caption{AUROC performance for predicting pregnancy across different race/ethnicity subgroups. Due to extremely small sample sizes, not all subgroups were reported.}
\label{table:preg_ethnicity}
\end{table}

\begin{table}[!htb]
\small
\centering
\begin{tabular}{lllll}
\toprule
Age & Baseline & WBM & PPG & WBM + PPG \\
\midrule
$<=$ 29 years (N (\%): 50,898 (1.6)) & 0.774 (0.756, 0.789) & 0.839 (0.825, 0.853) & 0.858 (0.843, 0.875) & \textbf{0.895 (0.882, 0.910)} \\
29 - 33 years (N (\%): 26,479 (2.5)) & 0.767 (0.751, 0.786) & 0.812 (0.798, 0.828) & 0.827 (0.808, 0.844) & \textbf{0.883 (0.869, 0.898)} \\
$>$ 33 years (N (\%): 102,840 (0.4)) & 0.884 (0.870, 0.895) & 0.867 (0.848, 0.888) & 0.862 (0.845, 0.878) & \textbf{0.925 (0.909, 0.939)} \\
\bottomrule
\end{tabular}
\caption{AUROC performance for predicting pregnancy across different age subgroups.}
\label{table:preg_age}
\end{table}

\begin{table}[!htb]
\centering
\scalebox{0.9}{
\begin{tabular}{lllll}
\toprule
Name & Baseline & WBM & PPG & WBM + PPG\\
\midrule
ACE-inhibitors & 0.792
(0.783, 0.800) & 0.799
(0.791, 0.808) & 0.825
(0.817, 0.833) & \textbf{\textit{0.829
(0.822, 0.837)}} \\
Active smoker & 0.777
(0.765, 0.789) & 0.843
(0.833, 0.853) & 0.859
(0.848, 0.868) & \textbf{\textit{0.881
(0.871, 0.890)}} \\
Afib & 0.799
(0.781, 0.814) & 0.847
(0.831, 0.862) & 0.826
(0.809, 0.842) & \textbf{\textit{0.860
(0.844, 0.874)}} \\
Allergy & 0.663
(0.656, 0.670) & 0.668
(0.661, 0.675) & 0.665
(0.659, 0.672) & \textbf{\textit{0.677
(0.670, 0.684)}} \\
Anti-anxiety & 0.693
(0.682, 0.704) & 0.729
(0.719, 0.739) & 0.748
(0.737, 0.758) & \textbf{\textit{0.761
(0.752, 0.772})} \\
Anti-psychotics & 0.795
(0.774, 0.814) & 0.821
(0.803, 0.839) & 0.822
(0.803, 0.841) & \textbf{\textit{0.832
(0.813, 0.851)}} \\
Anticoagulants & 0.710
(0.702, 0.718) & 0.768
(0.761, 0.776) & 0.856
(0.850, 0.863) & \textbf{0.866
(0.860, 0.872)} \\
Antidepressants & 0.787
(0.776, 0.800) & 0.793
(0.781, 0.805) & 0.794
(0.780, 0.806) & \textbf{\textit{0.800
(0.788, 0.812)}} \\
Antiplatelets & 0.767
(0.749, 0.783) & 0.830
(0.814, 0.845) & 0.855
(0.839, 0.868) & \textbf{\textit{0.873
(0.858, 0.885)}} \\
Anxiety & 0.754
(0.748, 0.760) & 0.785
(0.779, 0.790) & 0.806
(0.800, 0.811) & \textbf{\textit{0.822
(0.816, 0.827})} \\
Artery disease & 0.886
(0.863, 0.908) & \textbf{0.882
(0.852, 0.908)} & 0.881
(0.848, 0.909) & 0.872
(0.840, 0.901) \\
Arthritis & 0.790
(0.783, 0.797) & 0.794
(0.787, 0.800) & 0.787
(0.780, 0.793) & \textbf{\textit{0.801
(0.794, 0.808)}} \\
Asthma & 0.621
(0.612, 0.628) & 0.623
(0.614, 0.630) & 0.648
(0.640, 0.655) & \textbf{0.649
(0.641, 0.657)} \\
Beta-blockers & 0.800
(0.790, 0.811) & 0.825
(0.815, 0.835) & 0.789
(0.778, 0.800) & \textbf{\textit{0.844
(0.834, 0.854)}} \\
Blood pressure & 0.787
(0.780, 0.793) & 0.799
(0.793, 0.805) & 0.823
(0.818, 0.829) & \textbf{\textit{0.832
(0.826, 0.838)}} \\
Blood pressure med. & 0.703
(0.683, 0.722) & 0.720
(0.699, 0.738) & \textbf{0.744
(0.725, 0.762)} & \textbf{0.744
(0.725, 0.761)} \\
Calcium-channel Blockers & 0.793
(0.780, 0.806) & 0.804
(0.790, 0.817) & 0.825
(0.812, 0.839) & \textbf{\textit{0.858
(0.846, 0.869)}} \\
Cancer & \textbf{0.805
(0.793, 0.817)} & 0.792
(0.779, 0.804) & 0.804
(0.791, 0.816) & 0.800
(0.788, 0.812) \\
Chemotherapy & \textbf{0.742
(0.673, 0.797)} & 0.702
(0.633, 0.767) & 0.717
(0.641, 0.784) & 0.698
(0.621, 0.768) \\
Cholesterol & 0.753
(0.747, 0.759) & 0.758
(0.752, 0.765) & 0.768
(0.762, 0.775) & \textbf{\textit{0.774
(0.767, 0.780)}} \\
Chronic bronchitis & 0.730
(0.713, 0.746) & 0.750
(0.735, 0.766) & 0.748
(0.733, 0.764) & \textbf{0.759
(0.743, 0.774)} \\
Depression & 0.728
(0.722, 0.735) & 0.764
(0.758, 0.770) & 0.781
(0.775, 0.786) & \textbf{\textit{0.799
(0.794, 0.805)}} \\
Diabetes & 0.814
(0.803, 0.824) & 0.826
(0.816, 0.836) & 0.866
(0.855, 0.875) & \textbf{\textit{0.872
(0.862, 0.881})} \\
Diuretics & 0.735
(0.721, 0.748) & 0.748
(0.735, 0.761) & 0.771
(0.758, 0.782) & \textbf{0.775
(0.762, 0.787)} \\
Hearing & 0.728
(0.720, 0.736) & 0.725
(0.716, 0.734) & 0.723
(0.715, 0.732) & \textbf{\textit{0.734
(0.725, 0.742)}} \\
Heart attack & 0.846
(0.825, 0.865) & 0.858
(0.838, 0.876) & 0.846
(0.825, 0.865) & \textbf{0.866
(0.845, 0.885)} \\
Heart disease & 0.858
(0.842, 0.874) & 0.853
(0.837, 0.870) & 0.860
(0.844, 0.877) & \textbf{0.867
(0.849, 0.884)} \\
Heart failure & 0.824
(0.794, 0.850) & 0.858
(0.832, 0.881) & 0.865
(0.835, 0.890) & \textbf{\textit{0.889
(0.864, 0.910)}} \\
Heart rhythm & 0.664
(0.653, 0.676) & 0.682
(0.670, 0.694) & 0.689
(0.677, 0.700) & \textbf{\textit{0.703
(0.692, 0.715)}} \\
Hip/Knee & \textbf{0.855
(0.840, 0.870)} & 0.842
(0.825, 0.859) & 0.850
(0.835, 0.866) & 0.852
(0.836, 0.869) \\
Kidney & 0.698
(0.682, 0.713) & 0.708
(0.693, 0.723) & 0.707
(0.692, 0.722) & \textbf{0.714
(0.699, 0.729)} \\
Liver & 0.723
(0.675, 0.767) & 0.749
(0.698, 0.796) & 0.735
(0.691, 0.782) & \textbf{0.756
(0.709, 0.801)} \\
Lower back & 0.690
(0.681, 0.699) & 0.700
(0.691, 0.709) & 0.696
(0.687, 0.704) & \textbf{\textit{0.710
(0.701, 0.719)}} \\
Neck disorder & 0.732
(0.721, 0.742) & 0.738
(0.727, 0.749) & 0.736
(0.725, 0.748) & \textbf{\textit{0.747
(0.735, 0.758)}} \\
Neuropathy & 0.794
(0.780, 0.807) & 0.818
(0.805, 0.830) & 0.816
(0.803, 0.829) & \textbf{0.829
(0.815, 0.841)} \\
Opioid & 0.753
(0.733, 0.773) & 0.794
(0.774, 0.813) & 0.799
(0.780, 0.819) & \textit{0.820
(0.802, 0.839)} \\
On any medication & 0.735
(0.729, 0.742) & 0.763
(0.757, 0.769) & 0.796
(0.790, 0.801) & \textbf{\textit{0.806
(0.801, 0.811)}} \\
Osteoporosis & \textbf{0.872
(0.860, 0.885)} & 0.862
(0.847, 0.876) & 0.862
(0.846, 0.877) & 0.867
(0.851, 0.881) \\
Pacemaker & 0.821
(0.782, 0.857) & 0.858
(0.823, 0.891) & 0.897
(0.859, 0.931) & \textbf{0.899
(0.860, 0.933)} \\
Pain & 0.721
(0.703, 0.739) & 0.728
(0.711, 0.746) & 0.718
(0.702, 0.735) & \textbf{\textit{0.737
(0.721, 0.756)}} \\
Painkillers & 0.605
(0.596, 0.614) & 0.609
(0.600, 0.618) & 0.609
(0.600, 0.619) & \textbf{\textit{0.618
(0.609, 0.627)}} \\
Sleep apnea & 0.791
(0.784, 0.798) & 0.802
(0.795, 0.809) & 0.823
(0.815, 0.830) & \textbf{\textit{0.830
(0.823, 0.837)}} \\
Sleep medication & 0.642
(0.626, 0.655) & 0.689
(0.674, 0.702) & 0.700
(0.684, 0.714) & \textbf{\textit{0.718
(0.703, 0.731)}} \\
Stroke or TIA & 0.767
(0.742, 0.790) & 0.789
(0.765, 0.812) & 0.788
(0.764, 0.811) & \textbf{0.795
(0.770, 0.817)} \\
Thyroid & 0.757
(0.746, 0.768) & 0.754
(0.743, 0.765) & 0.756
(0.745, 0.766) & \textbf{0.761
(0.750, 0.772)} \\
Urinary & 0.805
(0.793, 0.818) & \textbf{0.814
(0.802, 0.827)} & 0.804
(0.790, 0.817) & 0.813
(0.800, 0.826) \\
Vision & 0.660
(0.652, 0.667) & 0.664
(0.657, 0.672) & 0.661
(0.653, 0.668) & \textbf{\textit{0.669
(0.662, 0.677)}} \\
\bottomrule
\end{tabular}}
\caption{AUROC performance metrics of different models across all baseline disease and medication tasks. The combination of both modalities results in the best performance for most tasks. Bold represents the best model(s) for each label, whereas italics represents results that are significantly better than the other approaches ($p < .05$)}
\label{table:hk_survey_full_results}
\end{table}

\end{document}